\documentclass{article} 
\usepackage{arxiv_stylefile,times}

\usepackage{amsmath,amsfonts,bm}

\def\eqref#1{equation~\ref{#1}}

\def\1{\bm{1}}

\def\rvd{{\mathbf{d}}}

\def\rvu{{\mathbf{i}}}

\def\rvu{{\mathbf{u}}}

\DeclareMathAlphabet{\mathsfit}{\encodingdefault}{\sfdefault}{m}{sl}
\SetMathAlphabet{\mathsfit}{bold}{\encodingdefault}{\sfdefault}{bx}{n}

\DeclareMathOperator*{\argmax}{arg\,max}

\usepackage{hyperref}
\usepackage{notation}
\usepackage{url}
\usepackage{enumitem}
\usepackage{algorithm}
\usepackage{algpseudocode}
\usepackage{subcaption}
\usepackage{multirow}
\usepackage{booktabs}

\usepackage{xcolor}
\definecolor{burgundy}{HTML}{7B2D26}

\title{When Models Don't Manipulate Manifolds: The Geometry of A Comparison Task}

\author{Sai Sumedh R. Hindupur$^1$, Hadas Orgad$^2$, Thomas Fel$^3$ and Demba Ba$^{1, 2}$  \\
School of Engineering and Applied Science, Harvard University$^1$, \\
Kempner Institute, Harvard University$^2$, Goodfire AI$^3$ \\
}

\iclrfinalcopy 
\begin{document}

\maketitle

\begin{abstract}
One of the current premises of mechanistic interpretability research is that detailed accounts of the geometry of neural network representations can tell us how models perform computations, and how to effectively intervene on them. While low dimensional manifolds have been observed for multiple concepts in the literature (e.g. numbers encoded on helices, days of the week on a circle, ...), with structure believed to reflect properties of data and tasks, the extent to which models rely on them for computation, and how they manipulate them, remains unclear. We characterize precisely the geometry of computation in a number-comparison task, as an abstraction of comparison for decision making, and how models utilize geometry in an elegant fashion to implement it. Specifically, we study the causal geometry of number comparison in Qwen2.5-7B-Instruct, a capable and widely studied open-weight model, and find Qwen largely uses linear representations of numbers despite the presence of curved geometry. To compare two numbers, the model first encodes each number along a vector and adds the two representations using attention and the residual connection, bringing them into a shared space in the residual stream. Then, the model uses MLP neurons to compare the pair of numbers on local regions in this shared space, which correspond to smaller intervals of input numbers, and combines these to obtain the position of the maximum. In fact, this reliance on linear representations for comparison also persists when the model compares three numbers. Our findings demonstrate that the manifold hypothesis can co-exist with linear representations: while concepts that are ordered may have manifold structure in representations, the model may use an underlying linear structure of the concept in certain computations. \footnote{Our code is available at \url{https://github.com/Sai-Sumedh/comparison-geometry}}
\end{abstract}

\section{Introduction}



How do language models represent concepts internally, and how do they use these representations to perform computations? The linear representation hypothesis \citep{park2023linear, elhage2021framework} posited that models represent ordered concepts as one-dimensional subspaces, which found evidence across diverse applications \citep{zhu2024languagemodelsrepresentbeliefs, marks2024geometrytruthemergentlinear, voynov2020unsuperviseddiscoveryinterpretabledirections, tigges2023linearrepresentationssentimentlarge, lee2024mechanisticunderstandingalignmentalgorithms}.  Recent work has challenged this perspective and found that these concepts live on low dimensional manifolds -- a phenomenon known as the manifold hypothesis: examples include number helices \citep{kantamneni2025languagemodelsusetrigonometry}, curved manifolds for dates/years \citep{modell2025originsrepresentationmanifoldslarge}, character count manifolds \citep{gurnee2026modelsmanipulatemanifoldsgeometry}, as well as age and temperature manifolds \citep{bhalla2026sparseautoencoderscaptureconcept}. Where does this manifold structure come from? Theoretical analyses suggest that the manifold structure arises from symmetries in data \citep{karkada2026symmetry} or task symmetries \citep{hwang2026intrinsic}. These insights have also revealed how steering along manifolds can control model behavior \citep{wurgaft2026manifoldsteeringrevealsshared}. However, given a specific computation, which aspects of the representation structure the model uses to perform the computation remains an open question. Does the model manipulate nonlinear concept manifolds, as \cite{gurnee2026modelsmanipulatemanifoldsgeometry} show using character count manifolds on a line-breaking task, or does the model use simpler linear representations for certain tasks despite the presence of manifold structure?

Elucidating how models manipulate internal representations to perform computations has become an active area of research.
Framed as such, this question  amounts to investigating the \textit{algorithmic} level in Marr's levels of analysis \citep{marr2010vision}, which claim that a system can be understood at the computational (abstract, behavioral), algorithmic (variables and how they are manipulated) and implementation levels (low-level implementation of the algorithm). \textit{Mechanistic interpretability} research, despite having largely focused on finding circuits of computation, namely the implementation level, has also become interested in the algorithmic level \citep{geiger2021causal} which operates at a higher level of abstraction. Multiple examples of algorithmic understanding of models exist, including trigonometry on circular/ helical structures for addition \citep{kantamneni2025languagemodelsusetrigonometry}, an addition mechanism using Fourier features for multiple concepts such as week days, months, etc  \citep{feucht2026arithmetic}, and explaining attention heads with python programs \citep{hayes2026explainingattentionprogramsynthesis}. Our work contributes to this burgeoning and exciting line of work on \textit{algorithmic interpretability}. 
\begin{figure}[t]
    \centering
    \includegraphics[width=\linewidth]{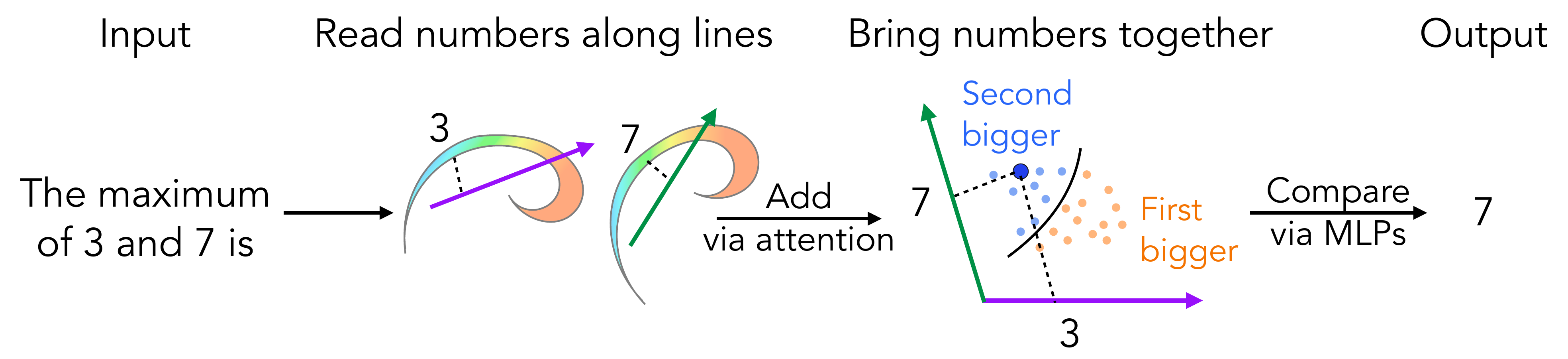}
    \caption{\textbf{Linear number representations are used by language model Qwen for comparison}. When asked to compare numbers, the model first represents them on a nonlinear manifold, but uses structure along a single linear direction in computation. It combines information about the two numbers by adding the corresponding linear representations using a single attention head and a residual connection, creating a shared space which enables comparison. On this shared space, the model identifies which number is larger using MLP neurons across two layers, and uses the answer position to read out the answer to the max task.}
    \label{fig:fig1toy}
    \vspace{-2em}
\end{figure}

Motivated by the above lines of inquiry, we ask how language models implement comparison using representation geometry. Comparison is a widely useful operation for intelligent systems, and language models in particular. It is an integral component of weighing options and making decisions, especially using numerical or ordered concepts (which can be represented on a line; see \citep{gardenfors2000conceptual}). For example, queries like \textit{"Which of these apartments is closest to my office?"}, \textit{“Which quarter had the worst sales in this five-year report?”} require performing comparison of the specified objects using the specified attributes. 
%
Despite reports of nonlinear, helical representations of numbers \citep{kantamneni2025languagemodelsusetrigonometry}, the aspects of number representations models use for the purpose of comparison, and how do they manipulate them to implement the operation remains poorly understood.

Previous works have studied how models perform comparison. \citep{hanna2023does} characterize in great detail a comparison circuit in GPT-2. However, they do not establish causal role of representation geometry in comparison by acknowledging that "GPT2’s structured number representations may be relevant to its greater-than ability. However, our
experiments struggle to prove this causally" (quoted from \cite{hanna2023does}). \citep{el-shangiti-etal-2025-geometry} found a linear subspace which causally affects model outputs, but their analysis does not concern the mechanism by which the model compares values represented in this subspace. \citep{yuchi2026llmsknownumberssay} study mixed notation number comparison and compare behavioral accuracy with classifier performance. Taken together, these works have focused on circuit discovery, model performance, or probing accuracy: they however do not study causal representation geometry and how it is used by the model to perform comparison. We provide a detailed geometric account of the algorithm language models employ to compare numbers (Fig. \ref{fig:fig1toy}). In addition to being descriptive, our account is causal: we can predict how causal interventions will affect model behavior.

Concretely, we make the following contributions:
\begin{itemize}[leftmargin=16pt, itemsep=1.5pt, topsep=1pt, parsep=1.5pt, partopsep=1pt]
    \item \textbf{Reconciling linear representations and manifolds}: We demonstrate how linear representations of numbers are causally involved in the model's number comparison, despite the presence of underlying curved manifold structure. 
    \item \textbf{Algorithm for number comparison using linear representations}: We further show how the model manipulates number representations to perform comparison: by additive mixing of individual number directions, followed by local comparisons (for specific intervals of input numbers) which are then combined to give the global comparison answer. 
    \item \textbf{Extension of the comparison algorithm to longer sequences of numbers}: We show how the algorithm using linear number representations extends to three-length sequences. 
\end{itemize}

\section{Pairwise Comparison of Numbers}

\begin{algorithm}[t]
\caption{Pairwise Number Comparisons} \label{alg:pairwise-compare}
\begin{algorithmic}
\Require{numbers $y_1, y_2$ at times $t=1, 2$ resp.}
\If{$t=1$}  
\State store $y_1$ as $\x_l^1 = \rvu f'_1(y_1)$ \Comment{Fig. \ref{fig:number-reps} (a, b)}
\ElsIf{$t=2$}
\State store $y_2$ as $\x_l^2 = \v_2 f_2(y_2)$ \Comment{Fig. \ref{fig:number-reps} (c)}
\State copy $\x_l^1$, transform  as $\v_1 f_1( y_1)$, \Comment{Done by Attention Head(s), Fig. \ref{fig:number-reps}(b, d)}
\State add to residual $\x_{l+1}^2 = \x_l^2 + \v_1 f_1( y_1)$ \Comment{Residual Stream, Fig. \ref{fig:shared-rep} (a--c)}
\State divide $\text{span}(\v_1, \v_2)$ into regions $\{\mathcal{R}_i\}$  \Comment{MLP neurons, Fig. \ref{fig:comparison-mechanism}(a, b)}
\State compare $y_1, y_2$ in $\{\mathcal{R}_i\}$ as $C_i = \mathbb{I}(y_1>y_2).\mathbb{I}(\x_{l+1}^2\in \mathcal{R}_i)$ \Comment{MLP neuron outputs, Fig. \ref{fig:comparison-mechanism}(b)}
\State combine local comparisons $\{C_i\}$ to get $ans = \mathbb{I}(y_1>y_2)$  \Comment{later MLP neurons, Fig. \ref{fig:comparison-mechanism}(c, d)}
\EndIf
\end{algorithmic}
\end{algorithm}

In this section, we first describe the pairwise comparison mechanism in an LLM. We state the algorithm explicitly, and discuss the main steps involved. Furthermore, we provide evidence describing how the model implements this algorithm in subsequent subsections. We extend the algorithm to three number comparisons and include evidence in Section \ref{sec:multi-compare}.

\paragraph{Notation.} Computationally relevant subspaces within model activations are denoted by $\x_l^t$, where $l$ denotes the layer index within the model and $t$ denotes time (token position). Numbers present in the input prompt are denoted by $\{y_t\}$. $\{\v_i\}$ are directions in model activations, which belong to the same space (same layer and token position). These directions encode numbers, with $\v_i$ encoding $\y_i$ as $\v_i f_i(y_i)$, where $f_i$ may be a nonlinear function of $y_i$. 

\subsection{Algorithm for Pairwise Comparisons}

In line with the known distinction between availability and utility of features \citep{garg2026featureslanguagemodelstore}, our claims in the following sections are about linear features for numbers being used by the model for a specific task: comparison. Other tasks and concepts, like addition/ periodic concepts \citep{feucht2026arithmetic, wurgaft2026manifoldsteeringrevealsshared}) may use more intricate manifold structure. 

The model encodes each number's magnitude $y_i$ as a nonlinear function $f_i(y_i)$ along a single direction $\v_i$, which is different for each number position. Since the two numbers $y_1, y_2$ are provided as inputs at different times (distinct token positions), the model then creates a \textit{shared} representation from the two numbers by (1) copying information about $y_1$ into $y_2$'s position, and (2) adding together the single number representations. The shared representation is expressed as:

\begin{align}
    \x = \v_1 f_1(y_1) + \v_2 f_2(y_2)
\end{align}

This representation spans a two-dimensional plane $\operatorname{span}(\v_1, \v_2)$. Moving along certain directions in this plane (e.g., along $\alpha \v_1 - \beta \v_2$ for any $\alpha>0, \beta\geq0$) changes the probability of the model answering $y_1$ as the greater number. The comparison $y_1>y_2$ can be linearly decoded on this plane. However, the model implements comparison in two stages, as described below. 

From this shared representation, MLP neurons first perform local comparisons, identifying and comparing the two numbers $y_1, y_2$ for specific ranges of individual numbers or their combinations. This occurs because of the gating-based nonlinearity of MLP neurons (SwiGLU for Qwen2.5-7B), whose sigmoid gate and overall expression leads to local regions of activation on intersection with $\text{span}(\v_1, \v_2)$. The outputs of the neurons are nonlinear on local regions, a consequence of approximate quadratic behavior of the SwiGLU nonlinearity on active regions.

Subsequent MLP neurons then combine these local comparisons to create global comparator neurons, which nearly perfectly capture $\mathbb{I}(y_1>y_2)$. These neurons then construct a single direction in the residual stream which encodes the comparison's answer, and causally affects the model's outputs. 

The algorithm is stated in Alg. \ref{alg:pairwise-compare}, visualized in App. Fig. \ref{fig:alg-flowchart}, along with evidence demonstrating each step.

\begin{figure}[t]
    \centering
    \includegraphics[width=\linewidth]{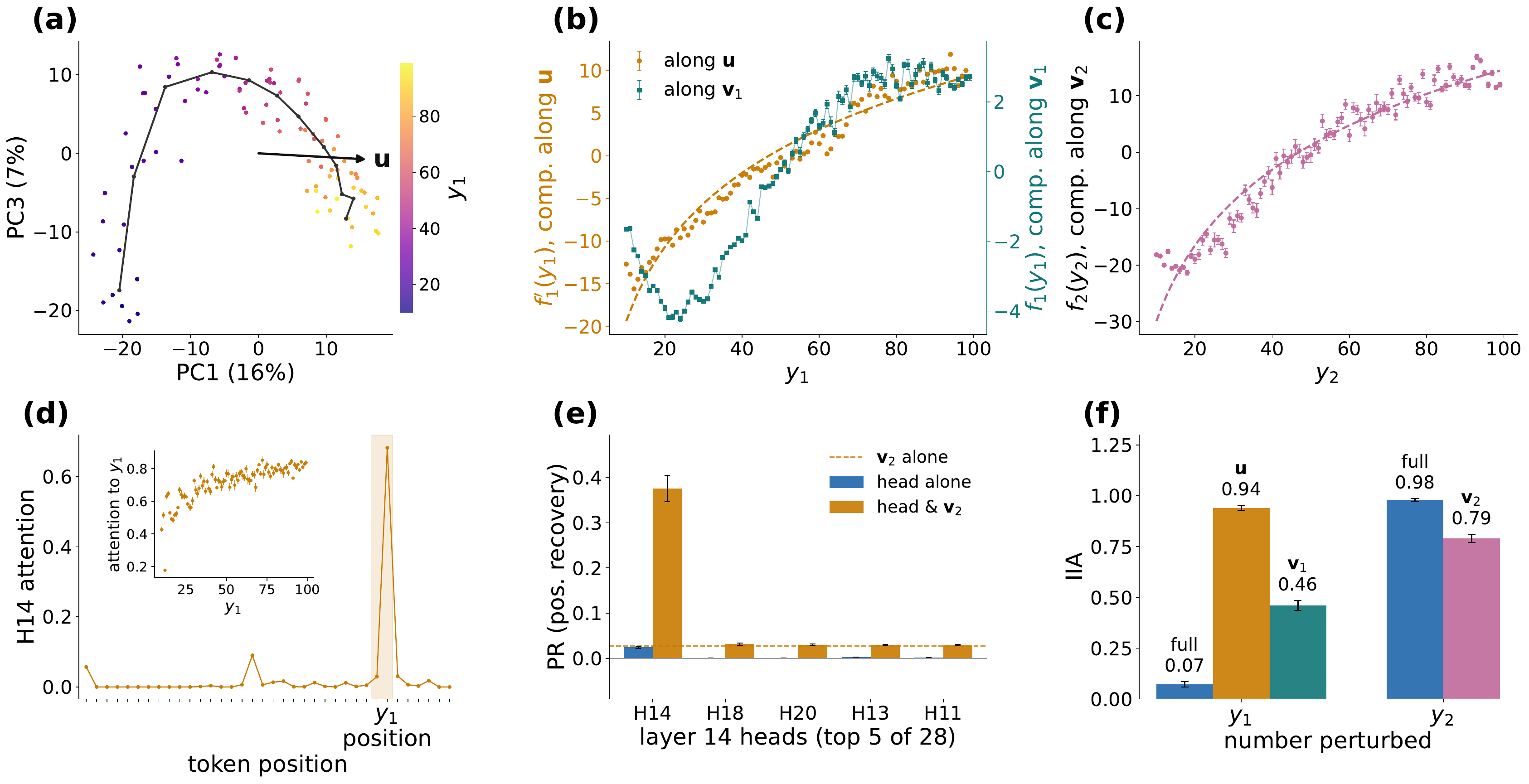}
    \caption{\textbf{A single causal direction for numbers controls model comparison.} \textbf{(a)} A causal direction $\rvu$ found in Layer 13's residual stream causally affects model behavior, despite the presence of curved geometry in the activations, as observed in principal components 1, 3. \textbf{(b, c)} Projection of activations onto the obtained causal directions $\rvu, \v_1, \v_2$ encode the number magnitude for a wide range of values. While $\rvu$ is in layer 13 residual stream at the first number $y_1$'s position, $\v_1, \v_2$ are causal directions encoding $y_1, y_2$ resp. at $y_2$'s position. $\v_1$ is obtained using DAS at an attention head H14's outputs. \textbf{(d)} Attention head 14 in layer 14 attends to the position of the first number $y_1$, irrespective of the number value, serving as a copy head. \textbf{(e)} Among all the attention heads in Layer 14, head H14 is causally involved in the model's computation, showing significantly higher position recovery than others. \textbf{(f)} Patching along $\rvu, \v_1, \v_2$ has significant causal effects on model behavior, as shown by interchange intervention accuracy (IIA).
    }
    \label{fig:number-reps}
    \vspace{-1.5em}
\end{figure}

\begin{figure}[t]
    \centering
    \includegraphics[width=\linewidth]{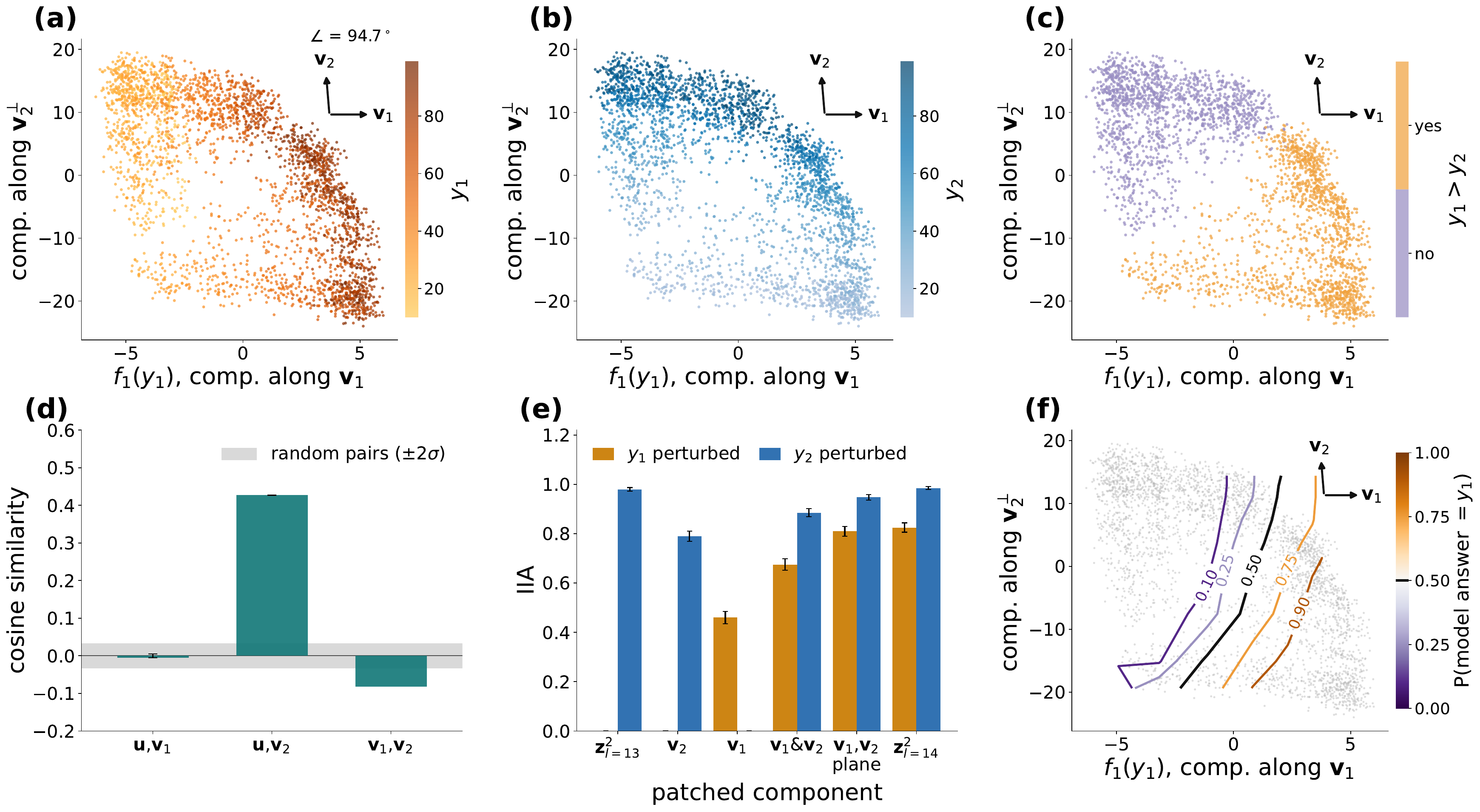}
    \caption{\textbf{The model uses causal number directions $\v_1, \v_2$ to construct a shared two-dimensional representation encoding the two inputs $y_1, y_2$. } \textbf{(a)}--\textbf{(c)} The linear span of $\v_1$ and $\v_2$ shows each number is encoded along its own direction, and the answer to comparison is linearly separable in this shared representation space. \textbf{(d)} By copying $y_1$ from $\rvu$ to $\v_1$, the model reduces the alignment between the two number representations: $|\cos(\v_1, \v_2)|<< |\cos(\rvu, \v_2)|$. \textbf{(e)} While the residual stream at layer 13, denoted $\z_{l=13}^2$, encodes $y_2$, $\v_1$ brings in causally useful information into the $y_2$ position. The shared representation is additive: Patching $\v_1 \& \v_2$ together (two rank-one patches) nearly matches the IIA of the rank-two patch onto the $\v_1, \v_2$ plane (i.e., $\text{span}(\v_1, \v_2)$). The plane itself captures as much information as the next layer 14 residual stream. \textbf{(f)} Moving around the $\v_1, \v_2$ plane, which is a two-dimensional plane in the 3,584-dimensional residual stream, is sufficient to change model behavior predictably.}  
    \label{fig:shared-rep}
\end{figure}

\textbf{Evidence for the pairwise comparison algorithm.} We perform experiments using the open-weight model Qwen2.5-7B-Instruct \citep{qwen25}. We ask the model to compare pairs of numbers, and provide the model with a one-shot example for output format. The prompt is "\textit{Answer in the following format with a single answer. The maximum of 12 and 4 is 12. The maximum of {y1} and {y2} is }". Our causal analyses involve activation patching using interchange interventions, where we patch specific component activations from a model running on a 'clean' prompt to when the model is processing another 'corrupt' prompt (\citep{meng2022rome}). For example, suppose the 'clean' prompt has inputs $(60, 10)$. The corrupt prompt then uses $(3, 10)$ as inputs, and patching activations from clean to corrupt changes the model's outputs on the 'corrupt' prompt. In this example, since the first number is changed between the clean and corrupt prompts, we refer to this as '$y_1$ perturbed' in subsequent figures Fig. \ref{fig:number-reps}, \ref{fig:shared-rep}, \ref{fig:comparison-mechanism}, \ref{fig:three-number-compare}. Further insights into our experimental setup is included in App. \ref{app:setup}. 

\textbf{The model computes the argmax position and uses that to produce the answer.} First, we observe that when patching is successful, patching model activations from a 'clean' run to a 'corrupted' run leads to changing the model's answer position, instead of the value (see App. \ref{app:res-behaviour}). For instance, patching activations from $(60, 10)$ into a model processing $(3, 10)$ will make the model answer $3$, and not $60$. Therefore, internal model activations compute the $\argmax$ position and use that to produce the answer. We restrict our analysis to the model's computation of the $\argmax$ position. 

We demonstrate the pairwise comparison algorithm by analyzing individual number representations (Fig. \ref{fig:number-reps}), which combine to form the shared representation (Fig. \ref{fig:shared-rep}). MLP neurons then perform the comparison in two stages, first locally and then globally (Fig. \ref{fig:comparison-mechanism}). 

\subsection{Individual number representations}

\textbf{Setup.} Using 2000 ordered pairs of two-digit numbers, which include 1000 unique randomly chosen pairs $(a,b)$ and their reflections $(b,a)$, we collect model activations at all layers and all token positions while processing the input prompt. The pairs $a,b$ are chosen to have distinct leading digits, so that the effect of patching is visible at the model logits since tokens are individual digits (two digit numbers having different first digits are a large fraction of all possible pairs, $\sim 90\%$). We use both PCA and Distributed Alignment Search (DAS) \citep{geiger2023das} to find causally relevant directions encoding each number $y_1, y_2$: we employ DAS whenever the principal components are not causally relevant. 

\textbf{Observations.} Fig. \ref{fig:number-reps}a shows that despite nonlinear manifold structure of number representations (shown in PC1-PC3 projection), there exists a direction $\rvu$ in layer 13's residual stream which causally affects the model's answer, as measured by Interchange Intervention Accuracy (IIA, \citep{geiger2021causal}) (Fig. \ref{fig:number-reps}f). Note that we compute IIA using the immediate next token generated by the model. However, the IIA scores are very similar for patching along directions and subspaces of interest even when computed using the entire number generated by the model (see App. \ref{app:res-whole}). A single attention head H14 of layer 14 copies information about $y_1$ from layer 13 in token position $y_1$ to $y_2$. It consistently attends to the $y_1$ position (Fig. \ref{fig:number-reps} d), and patching this head's output has the highest effect on the model's logits, as measured using position recovery (a modified version of recovery \citep{meng2022rome} that patching changes the model's answer position instead of value).
\begin{align}
\label{eq:position-recovery}
    PR = \frac{LD_{patch} - LD_{corrupt}}{LD_{clean}-LD_{corrupt}}
\end{align}
where LD is the logit difference between the pair $(\min(y_1, y_2), \max(y_1, y_2))$ where $y_1, y_2$ are inputs on the corrupt prompt, since upon patching from clean to corrupt prompts, models output $\min(y_1, y_2)$ as the answer to the corrupt max-prompt. Note that we call this position recovery to observe how well the model reorganizes its logits to patching and answers with the patched position. The denominator is only meant to provide a rough scale of logit difference.
$\v_1$ is then obtained as the DAS direction at the output of H14. $\v_2$ is obtained as the top principal component from the layer 13 residual stream. Therefore, obtained causal directions $\rvu, \v_1, \v_2$ encode the values of $y_1, y_1, y_2$ respectively, as shown in panels (b, c). These directions are causal, as shown by their IIA scores in (Fig. \ref{fig:number-reps}f).

\subsection{Shared Representation enables comparison}

\begin{figure}[t]
    \centering
    \includegraphics[width=\linewidth]{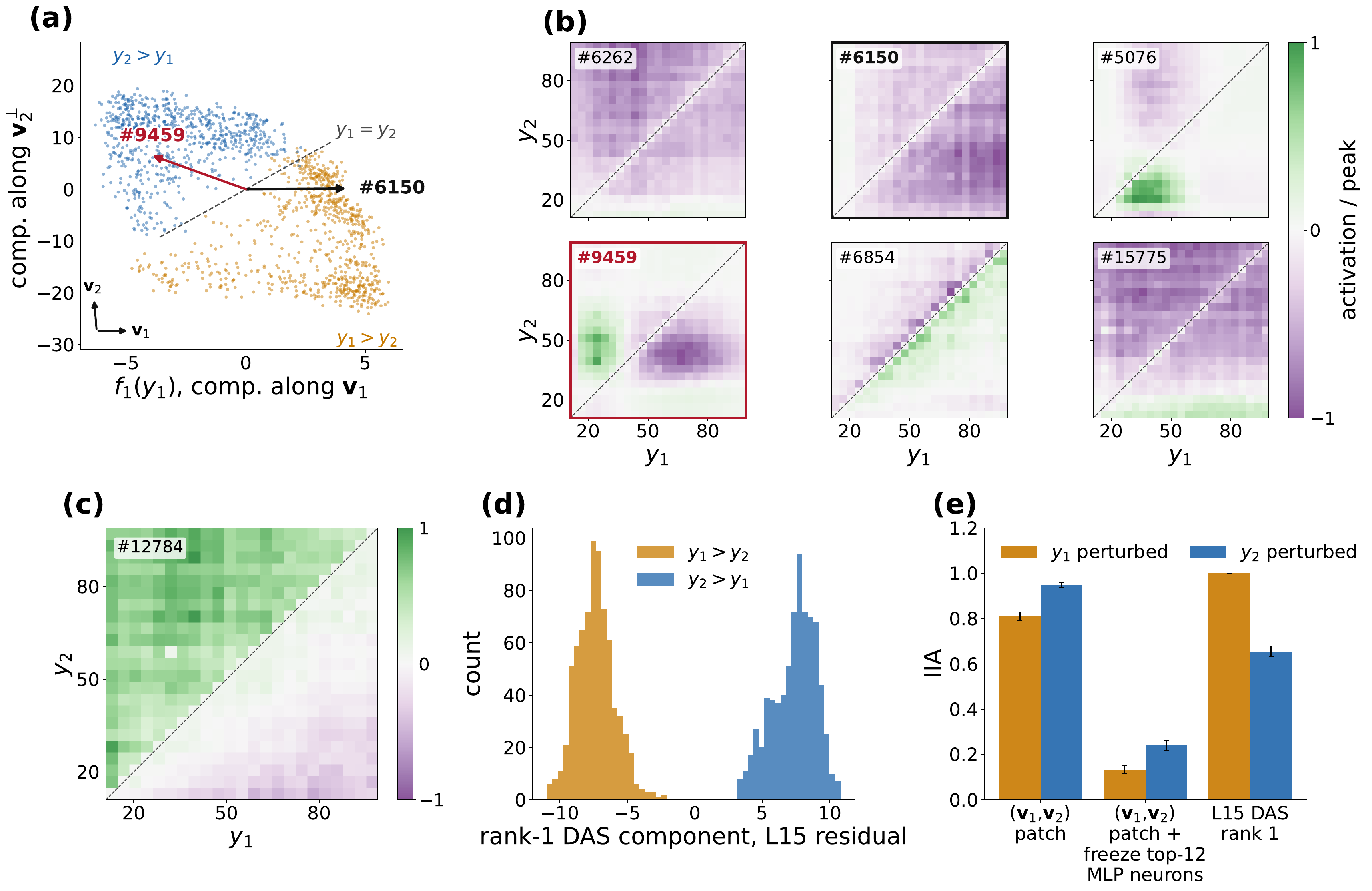}
    \caption{\textbf{The model compares numbers $y_1, y_2$ by combining local comparisons on the shared $\v_1, \v_2$ plane}. \textbf{(a)} Individual neuron weights in layer 14 MLP are specific directions in the $\v_1, \v_2$ plane. The y-axis is $\v_2^\perp$, the component of $\v_2$ orthogonal to $\v_1$ (since $\v_1, \v_2$ are not exactly orthogonal) \textbf{(b)} Neurons in layer 14 MLP (ranked by attribution scores) localize specific regions of the inputs $y_1, y_2$ (like neuron $\# 6150$), or perform comparisons in localized regions (neuron $\#9459$). \textbf{(c)} Select neurons in layer 15 MLP, which combine the outputs of layer 14 MLP neurons, are global comparators: they respond positively when $y_2>y_1$ and negatively otherwise. \textbf{(d)} A single direction in the layer 15 residual stream (after layer 15 MLP), which is formed by inputs from global comparator neurons from layer 15 MLP, encodes the position of the answer $\argmax (y_1, y_2)$. \textbf{(e)} There are 12 comparator neurons (6 each in layer 14, 15  MLPs) which perform comparison: freezing these neurons significantly degrades the IIA achieved by patching in the $(v_1, v_2)$ plane . The one-dimensional comparison direction in layer 15 (panel (d)) controls model behavior.}
    \label{fig:comparison-mechanism}
    \vspace{-2em}
\end{figure}

Using the causal directions of individual numbers $\v_1, \v_2$, the model creates a shared representation of both numbers, which is a two-dimensional plane in the pre-MLP residual stream of layer 14. We visualize the projection of model activations onto the linear span of these directions (Fig. \ref{fig:shared-rep}). This shared representation lives in the pre-MLP residual stream of layer 14. Fig. \ref{fig:shared-rep} (a)--(c) show activations in this space encoding each number along its own direction, while allowing linear separation of the two cases $y_1>y_2, y_1<y_2$. The probability contours of the model's answer are approximately parallel to the decision boundary (Fig. \ref{fig:shared-rep} f). The directions $\v_1, \v_2$ are nearly orthogonal (Fig. \ref{fig:shared-rep} d) and less aligned than their counterparts which lived at different token positions ($\rvu, \v_2$). Patching these directions using two rank-one patches (i.e., by projecting along $\v_1$ and $\v_2$) shows a similar causal effect (IIA) as patching the rank-two shared representation space (by projecting onto $\text{span}(\v_1, \v_2)$) (Fig. \ref{fig:shared-rep} e), indicating linear combination of individual number representations create this shared space. This observation is nontrivial because $\v_2$ is obtained in layer 13's residual stream and $\v_1$ is from head H14's output (which itself uses the layer 13 residual stream in computation), making nonlinear interactions possible.

\vspace{-1em}
\subsection{Global Comparison is a combination of local comparisons}

\textbf{Setup.} We first identify which layer MLPs are involved in the maximum computation by performing freezing experiments. Here, we patch the $\text{span}(\v_1, \v_2)$ space, but freeze the downstream MLP outputs to remain the same as the no-patch case. This allows us to check the contribution of individual layer MLPs in the subsequent computation: we expect that freezing important MLPs will result in a significant drop in the patching effectiveness.
For relevant MLPs, we then identify neurons of interest using first-order attribution scores from the model logits (App. \ref{app:methods-attribution}, App. \ref{app:res-neurons}). The obtained neurons are tested for causal relevance by further patch-and-freeze experiments (see App. \ref{app:methods} for details).

\textbf{Observations.} MLP neurons in layers 14 and 15 operate on the shared representation space $\text{span}(\v_1, \v_2)$ (Fig. \ref{fig:comparison-mechanism}(a)). We find a set of 12 neurons (6 neurons in MLP of layer 14 and 6 in layer 15's MLP) which together contribute to the comparison computation, and freezing these neurons significantly reduces the IIA from patching the $(\v_1, \v_2)$ space (see Fig. \ref{fig:comparison-mechanism} (e) and App. \ref{app:res-neurons}). They perform the comparison in two stages: first, neurons in the MLP of layer 14 perform localized comparisons: they respond to specific ranges of values of $y_1$ or $y_2$, or a combination thereof (e.g., $|y_1-y_2|<\eta$) (see Fig. \ref{fig:comparison-mechanism} (b) which shows the receptive fields of neurons, i.e., their activations as a heatmap in the $(y_1, y_2)$ space). While the geometry of the $\text{span}(\v_1, \v_2)$ space makes comparison possible by linear readouts (Fig. \ref{fig:shared-rep}c), MLP neurons seem to use their linear transform followed by nonlinearity to perform local comparisons. Some neurons perform the comparison $y_1<y_2$ on these bounded ranges (like neuron $\# 9459$ in Fig.\ref{fig:comparison-mechanism}(b)) while others serve to identify the ranges (like neuron $\# 6150$ in Fig. \ref{fig:comparison-mechanism}(b)). These neurons are then combined in the second stage (MLP of layer 15), into neurons which serve as global comparators, encoding the answer position over the entire range of values of both numbers (Fig. \ref{fig:comparison-mechanism}(c)). We find a single direction (using DAS) in layer 15's residual stream (after MLP of layer 15), which is written to by several global comparator neurons, and encodes the answer to the comparison as a binary value (Fig.\ref{fig:comparison-mechanism}(d)). The importance of these twelve MLP neurons, as well as the causal relevance of the DAS direction obtained in layer 15, are shown using IIA in panel (Fig. \ref{fig:comparison-mechanism}(e)).

\section{Multiple Comparisons: The Case with Three Numbers}
\label{sec:multi-compare}

\begin{algorithm}[t]
\caption{Three-Number Comparisons} \label{alg:multi-compare}
\begin{algorithmic}
\Require{numbers $y_1, y_2, y_3$ at times $t=1, 2, 3$ resp., $\Tilde{T}$ is the last token.}
\If{$t=1$}  
\State store $y_1$ as $\x_l^1 = \rvu_1 f_1(y_1)$
\ElsIf{$t=2$}
\State store $y_2$ as $\x_l^2 = \rvu_2 f_2(y_2)$
\State compute $\alpha_2 = \mathbb{I}(y_2=\max(y_1, y_2))$ and store along $\rvd_2$
\Comment{Use Alg. \ref{alg:pairwise-compare}; Fig. \ref{fig:three-number-compare} (d), App. Fig. \ref{fig:app-k3-summary}}
\ElsIf{$t=3$}
\State store $y_3$ as $\x_l(3) = \v_3 f_3(y_3)$ \Comment{Fig. \ref{fig:three-number-compare}(a)}
\State copy $\x_l^1, \x_l^2$, transform as $\v_1 f_1(y_1), \v_2 f_2(y_2)$ \Comment{Done by Attention Heads, App. Fig. \ref{fig:app-k3-summary}}
\State add to residual $\x_{l+1}(3) = \x_l(3) +\v_1 f_1(y_1) + \v_2 f_2(y_2)$ \Comment{Residual Stream, App. Fig. \ref{fig:app-k3-summary}}
\State divide $\text{span}(\v_1, \v_2, \v_3)$ into regions $\mathcal{R}_i$ \Comment{Fig. \ref{fig:three-number-compare}(c)}
\State compare in $\mathcal{R}_i$ as $C_i = \mathbb{I}(y_3=\max(y_1, y_2, y_3)).\mathbb{I}(\x_{l+1}(3) \in \mathcal{R}_i)$ \Comment{MLP neurons, Fig. \ref{fig:three-number-compare}(c)}
\State combine $\{C_i\}$, get $\alpha_3 = \mathbb{I}(y_3 = \max(y_1, y_2, y_3))$, store on $\rvd_3$ \Comment{Later MLP neurons, Fig. \ref{fig:three-number-compare}(d)}
\ElsIf{$t=\Tilde{T}$}
\State combine $\alpha_2 \rvd_2, \alpha_3 \rvd_3$ to get $\x_{\ell'}(\Tilde{T})=\sum_t w_t \alpha_t \rvd_t$  \Comment{Attention Head, Fig. \ref{fig:three-number-compare}(e), App. Fig. \ref{fig:app-k3-summary}}
\State Read $\argmax(y_1, y_2, y_3)$ from $\x_{\ell'}(\Tilde{T})$
\EndIf
\end{algorithmic}
\end{algorithm}

Having stated and established evidence for the pairwise comparisons algorithm in Qwen, we now extend it to multiple number comparisons: does the model continue to use linear representations of numbers for multiple comparisons? The model is very good at the multiple comparison task: comparing long sequences of numbers in a single forward pass (see App. \ref{app:res-behaviour}). Using three numbers as a case study, we state the model's algorithm and describe evidence for this algorithm in model activations.

\begin{figure}[t]
    \centering
    \includegraphics[width=\linewidth]{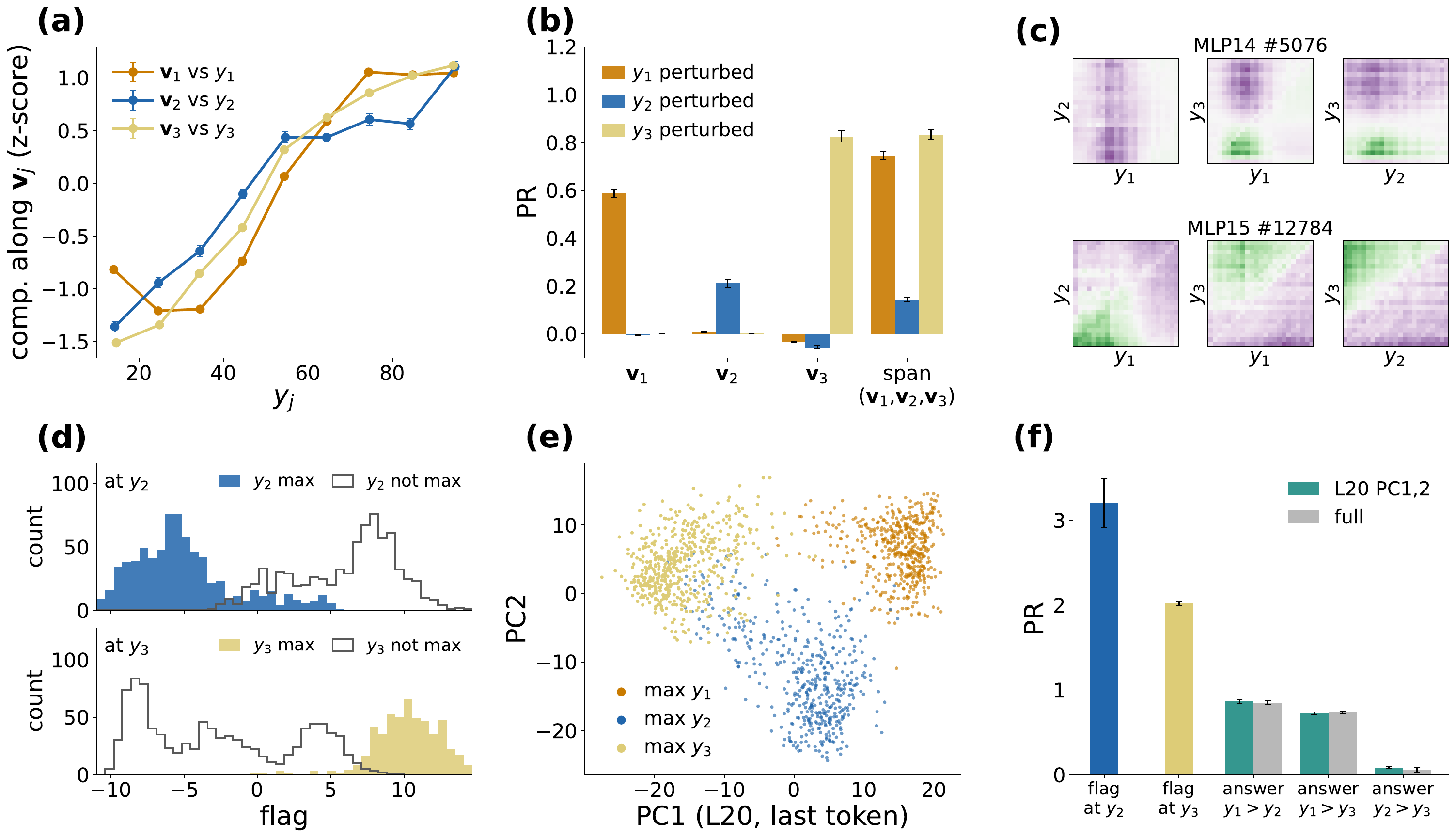}
    \caption{\textbf{The model continues to use linear number representations for three-number comparisons.} \textbf{(a)} We observe that the causal directions $\v_1, \v_2, \v_3$ (obtained at the third number $y_3$'s position) encode the numbers $y_1, y_2, y_3$, through nearly monotonic components $f_i(y_i)$. \textbf{(b)} The three directions $\v_1, \v_2, \v_3$ have causal effects on the model outputs (logits, measured using position recovery, PR). The effect on $y_2$ is lower at this position ($y_3$ position). \textbf{(c)} Receptive fields from layer 14, layer 15 MLP neurons show local comparisons now being performed in the $y_1, y_2, y_3$ space (whose three two-dimensional views are shown). \textbf{(d)} The model represents its answer \textit{flag} along a single direction in the layer 15 residual stream at both the $y_2$ and the $y_3$ position. The flag identifies if $y_t$ is the largest upto time $t$. Note that $y_2 \max$ refers to $y_2>y_1$ here. \textbf{(e)} The $\argmax$ position is represented in the top two principal components of layer 20 residual stream, at the last token position (after the third number $y_3$, and right before the model answers). \textbf{(f)}. The flags at $y_2$ and $y_3$ positions, as well as the answer position from (e), are causal and affect model logits.}
    \label{fig:three-number-compare}
    \vspace{-1em}
\end{figure}

\subsection{Algorithm for Multiple Comparisons}

To compare longer sequences of numbers, language models employ an interesting strategy: at every time $t$, they compute a binary flag which answers the following question: is $y_t$ the largest number seen so far? To achieve this, they construct an additive sum of linear representations of all numbers up to and including $y_t$. Individual neurons then perform local comparisons on this shared representation, which is combined to give a global comparator that identifies whether $y_t$ is the largest entry seen so far. This information is stored along a single direction $\rvd_t$, as a binary variable $\alpha_t = \mathbb{I}(y_t=\max (y_1, \dots, y_{t}))$. 
\begin{equation}
\begin{aligned}
    \x &= \v_1 f_1(y_1) + \v_2 f_2(y_2) + \v_3 f_3(y_3), \\
    \text{at } t,\quad \alpha_t
       &= \mathbb{I}\!\left(y_t = \max(y_1, \dots, y_{t})\right).
\end{aligned}
\end{equation}

At this stage of computation, if $y_t$ is not the maximum upto time $t$, i.e., the flag $\alpha_t=0$, the position of the maximum is not otherwise stored (i.e., the model simply knows that the maximum is not at $t$, but not if the maximum is at an intermediate position). Information about whether the first number $y_1$ (which is a special case since there are no numbers before it) is the maximum is also computed at the $y_3$ position. At the final token $\Tilde{T}$ before generating the answer, the model combines these directions using a single attention head which attends to the answer position (i.e., $w_t$ is high at $\argmax(y_1, \dots, y_T)$) to obtain $\x_{\ell'}(\Tilde{T}) = \sum_{t=1}^T w_t \alpha_t \rvd_t$ (App. Fig. \ref{fig:app-k3-summary}). This final representation encodes the answer position which is then read off by the model. While our results describe how individual number representations are used to create causally relevant flags, we empirically find the above attention head. We leave an investigation into the complete mechanism of how the head works, and how the model decodes the answer position to future work.

We hypothesize that the model may read this information by exploiting the order of numbers, checking sequentially (in decreasing $t$) if $\alpha_t=1$ and identifying the first time this occurs. We state the algorithm in Alg. \ref{alg:multi-compare}.

\subsection{Evidence for the three number comparison algorithm}

\textbf{Setup.} We extend our analysis to studying comparison of three numbers. We choose $~1500$ triples and $~400$ held-out examples. In this case, IIA does not show enough signal for analysis, so we employ position recovery (Eq. \ref{eq:position-recovery}) to test for causal effects of activation patching on the model's logits.

\textbf{Observations.} First, we note that when given three numbers, the model seems to perform two parallel computations (see App. \ref{app:res-three}). 
Fig. \ref{fig:three-number-compare}'s first row shows the computation occurring at the third number position. Using three directions $\v_1, \v_2, \v_3$ which encode the magnitudes of the three numbers $y_1, y_2, y_3$ (panel a), the model creates a three-dimensional space $\text{span}(\v_1, \v_2, \v_3)$. Fig. \ref{fig:three-number-compare}(b) shows that patching each direction affects the model outputs when the corresponding number is perturbed, and that patching the span affects multiple numbers. The effect of perturbing $y_2$ is minimal here, indicating that the model uses different computation for determining $y_2$ is the largest. Representative MLP neurons in layer 14 MLP perform local comparisons in the $(y_1, y_2, y_3)$ space (Fig. \ref{fig:three-number-compare}(c)). These neurons construct a single direction in layer 15's residual stream space which encodes the answer $\mathbb{I}(y_3=\max(y_1, y_2, y_3))$ (Fig. \ref{fig:three-number-compare}(d), bottom). At the second number $y_2$'s position, the model compares $y_2$ and $y_1$ (using Alg. \ref{alg:pairwise-compare}), and creates a single direction from DAS which encodes the answer position in layer 15's residual stream (Fig. \ref{fig:three-number-compare}(d), top). 
At the last token position, layer 20's residual stream combines the flag information (using an attention head, see App. Fig. \ref{fig:app-k3-summary}) encodes the position of the answer, as visible on a two dimensional PCA Fig. \ref{fig:three-number-compare}(e). Position recovery scores in Fig. \ref{fig:three-number-compare}(f) show that the flags at the $y_2$ and $y_3$ positions, as well as the two-dimensional subspace encoding the overall answer in layer 20 (from Fig. \ref{fig:three-number-compare}(e)) are causal. The two principal components in layer 20 have the same position recovery as the entire residual stream at that position (Fig. \ref{fig:three-number-compare}(f)). 

\vspace{-1em}
\section{Discussion and Limitations}
Our work is a concrete illustration of how a language model (Qwen2.5-7B-Instruct in our case) uses linear representations and additive mixing for specific model computations, number comparison in our case. The model uses linear representations for each number position despite the presence of a nonlinear manifold in its activations. We demonstrate how the model methodically combines information from both numbers by adding the number representations together, constructing a two-dimensional plane in its residual stream. The model then employs MLP neurons to focus on specific local regions within this space, which correspond to smaller intervals of the input numbers, and performs comparison on these patches. Finally, the model combines these local patches into a global comparison. In addition, we show how this algorithm extends to three number comparisons. Our findings challenge a central hypothesis in current mechanistic interpretability research, namely that detailed accounts of the representation geometry of individual concepts necessarily make it easier to characterize how models perform meaningful computations using these concepts. As we demonstrate, even though one can precisely characterize the geometry of number manifolds in model activations, simpler linear representations of numbers may be enough to causally affect the model's computation. Our findings complement and add nuance to our understanding of what representation geometry of neural networks truly teaches us: while the intricacies of their geometric structure may reflect the properties of the underlying data distribution, simpler structures within this geometry may be sufficient computationally and useful to the model.
\newpage
\textbf{Limitations.} The limitations and assumptions of our work are stated below:
\begin{itemize}[leftmargin=16pt, itemsep=1.5pt, topsep=1pt, parsep=1.5pt, partopsep=1pt]
    \item Our analysis is restricted to a specific kind of input -- numbers -- and a specific operation -- computing the maximum. Our claims about linear representations being useful despite the presence of manifold geometry hold for concepts that are ordered, i.e., those that can be mapped to the number line (without periodicity).
    \item We rely on DAS \citep{geiger2023das} to obtain causal directions in model activations, whenever principal components have low causal effects. Therefore, our analysis may partially inherit the same challenges as DAS, such as finding shortcuts \citep{wu2024interpretabilityscaleidentifyingcausal}. However, we include additional evidence grounded in the model, such as neurons, attention heads, etc which may partially alleviate concerns of shortcuts.
    \item Much of our analysis shows sufficiency: we identify directions and subspaces which can modify the model's outputs when patched. We don't claim necessity: removing these components may not affect the model's outputs -- the model can still use other paths of information processing within its layers to solve the same task (like the hydra effect \citep{mcgrath2023hydraeffectemergentselfrepair}).
\end{itemize}

\subsubsection*{Acknowledgments}
This work has been made possible in part by a gift from the Chan Zuckerberg Initiative Foundation to establish the Kempner Institute at Harvard University. SSRH and DB thank the Kempner Institute for access to compute resources. SSRH thanks members of the CRISP lab at Harvard SEAS for useful discussions and feedback on the manuscript. SSRH further thanks Atticus Geiger and Ekdeep Singh Lubana for insightful discussions about the project. 
\subsection*{AI use statement}

In this work, we used generative AI tools for writing and editing code, literature search, drafting portions of the appendix, and feedback on research ideas and experimental methodology. We have reviewed and verified all AI-assisted work. We take responsibility for the final content of this work, including text, claims, code, or artifacts produced with the aid of generative AI.

\subsection*{Reproducibility statement}
Appendix \ref{app:setup} provides the experimental setup, prompts, datasets and sampling procedures, while Appendix \ref{app:methods} provides details of the intervention, DAS, attribution, and evaluation methods. Sample sizes and hyperparameters are summarized in Table \ref{app:setup-hparams}.




\bibliography{arxiv/ref_arxiv}

\begin{thebibliography}{40}
\providecommand{\natexlab}[1]{#1}
\providecommand{\url}[1]{\texttt{#1}}
\expandafter\ifx\csname urlstyle\endcsname\relax
  \providecommand{\doi}[1]{doi: #1}\else
  \providecommand{\doi}{doi: \begingroup \urlstyle{rm}\Url}\fi

\bibitem[Alain \& Bengio(2016)Alain and Bengio]{alain2016probes}
Guillaume Alain and Yoshua Bengio.
\newblock Understanding intermediate layers using linear classifier probes.
\newblock \emph{arXiv preprint arXiv:1610.01644}, 2016.

\bibitem[Belinkov(2022)]{belinkov2022survey}
Yonatan Belinkov.
\newblock Probing classifiers: Promises, shortcomings, and advances.
\newblock \emph{Computational Linguistics}, 48\penalty0 (1):\penalty0 207--219, 2022.
\newblock \doi{10.1162/coli_a_00422}.

\bibitem[Bhalla et~al.(2026)Bhalla, Fel, Rager, Feucht, Haklay, Wurgaft, Boppana, Kowal, Shyam, Merullo, Geiger, and Lubana]{bhalla2026sparseautoencoderscaptureconcept}
Usha Bhalla, Thomas Fel, Can Rager, Sheridan Feucht, Tal Haklay, Daniel Wurgaft, Siddharth Boppana, Matthew Kowal, Vasudev Shyam, Jack Merullo, Atticus Geiger, and Ekdeep~Singh Lubana.
\newblock Do sparse autoencoders capture concept manifolds?, 2026.
\newblock URL \url{https://arxiv.org/abs/2604.28119}.

\bibitem[Brown et~al.(2001)Brown, Cai, and DasGupta]{brown2001interval}
Lawrence~D. Brown, T.~Tony Cai, and Anirban DasGupta.
\newblock Interval estimation for a binomial proportion.
\newblock \emph{Statistical Science}, 16\penalty0 (2):\penalty0 101--133, 2001.
\newblock \doi{10.1214/ss/1009213286}.

\bibitem[Chan et~al.(2022)Chan, Garriga-Alonso, Goldowsky-Dill, Greenblatt, Nitishinskaya, Radhakrishnan, Shlegeris, and Thomas]{chan2022scrubbing}
Lawrence Chan, Adri{\`a} Garriga-Alonso, Nicholas Goldowsky-Dill, Ryan Greenblatt, Jenny Nitishinskaya, Ansh Radhakrishnan, Buck Shlegeris, and Nate Thomas.
\newblock Causal scrubbing: A method for rigorously testing interpretability hypotheses.
\newblock AI Alignment Forum, \url{https://www.alignmentforum.org/posts/JvZhhzycHu2Yd57RN/causal-scrubbing-a-method-for-rigorously-testing}, 2022.

\bibitem[El-Shangiti et~al.(2025)El-Shangiti, Hiraoka, AlQuabeh, Heinzerling, and Inui]{el-shangiti-etal-2025-geometry}
Ahmed~Oumar El-Shangiti, Tatsuya Hiraoka, Hilal AlQuabeh, Benjamin Heinzerling, and Kentaro Inui.
\newblock The geometry of numerical reasoning: Language models compare numeric properties in linear subspaces.
\newblock In Luis Chiruzzo, Alan Ritter, and Lu~Wang (eds.), \emph{Proceedings of the 2025 Conference of the Nations of the Americas Chapter of the Association for Computational Linguistics: Human Language Technologies (Volume 2: Short Papers)}, pp.\  550--561, Albuquerque, New Mexico, April 2025. Association for Computational Linguistics.
\newblock ISBN 979-8-89176-190-2.
\newblock \doi{10.18653/v1/2025.naacl-short.47}.
\newblock URL \url{https://aclanthology.org/2025.naacl-short.47/}.

\bibitem[Elhage et~al.(2021)Elhage, Nanda, Olsson, Henighan, Joseph, Mann, Askell, Bai, Chen, Conerly, et~al.]{elhage2021framework}
Nelson Elhage, Neel Nanda, Catherine Olsson, Tom Henighan, Nicholas Joseph, Ben Mann, Amanda Askell, Yuntao Bai, Anna Chen, Tom Conerly, et~al.
\newblock A mathematical framework for transformer circuits.
\newblock \emph{Transformer Circuits Thread}, 2021.
\newblock URL \url{https://transformer-circuits.pub/2021/framework/index.html}.

\bibitem[Feucht et~al.(2026)Feucht, Haklay, Bhalla, Wurgaft, Rager, Sarfati, Merullo, McGrath, Lewis, Lubana, et~al.]{feucht2026arithmetic}
Sheridan Feucht, Tal Haklay, Usha Bhalla, Daniel Wurgaft, Can Rager, Rapha{\"e}l Sarfati, Jack Merullo, Thomas McGrath, Owen Lewis, Ekdeep~Singh Lubana, et~al.
\newblock Arithmetic in the wild: Llama uses base-10 addition to reason about cyclic concepts.
\newblock \emph{arXiv preprint arXiv:2605.01148}, 2026.

\bibitem[G{\"a}rdenfors(2000)]{gardenfors2000conceptual}
Peter G{\"a}rdenfors.
\newblock \emph{Conceptual spaces}, volume~3.
\newblock MIT press Cambridge, MA, 2000.

\bibitem[Garg et~al.(2026)Garg, Kleinberg, and Peng]{garg2026featureslanguagemodelstore}
Nikhil Garg, Jon Kleinberg, and Kenny Peng.
\newblock How many features can a language model store under the linear representation hypothesis?, 2026.
\newblock URL \url{https://arxiv.org/abs/2602.11246}.

\bibitem[Geiger et~al.(2021)Geiger, Lu, Icard, and Potts]{geiger2021causal}
Atticus Geiger, Hanson Lu, Thomas Icard, and Christopher Potts.
\newblock Causal abstractions of neural networks.
\newblock In \emph{Advances in Neural Information Processing Systems (NeurIPS)}, volume~34, pp.\  9574--9586, 2021.
\newblock arXiv:2106.02997.

\bibitem[Geiger et~al.(2024)Geiger, Wu, Potts, Icard, and Goodman]{geiger2023das}
Atticus Geiger, Zhengxuan Wu, Christopher Potts, Thomas Icard, and Noah~D. Goodman.
\newblock Finding alignments between interpretable causal variables and distributed neural representations.
\newblock In \emph{Proceedings of the Third Conference on Causal Learning and Reasoning (CLeaR)}, volume 236 of \emph{Proceedings of Machine Learning Research}, pp.\  160--187. PMLR, 2024.
\newblock arXiv:2303.02536.

\bibitem[Goldowsky-Dill et~al.(2023)Goldowsky-Dill, MacLeod, Sato, and Arora]{goldowskydill2023pathpatching}
Nicholas Goldowsky-Dill, Chris MacLeod, Lucas Sato, and Aryaman Arora.
\newblock Localizing model behavior with path patching.
\newblock \emph{arXiv preprint arXiv:2304.05969}, 2023.

\bibitem[Gurnee \& Tegmark(2024)Gurnee and Tegmark]{gurnee2023space}
Wes Gurnee and Max Tegmark.
\newblock Language models represent space and time.
\newblock In \emph{International Conference on Learning Representations (ICLR)}, 2024.
\newblock arXiv:2310.02207.

\bibitem[Gurnee et~al.(2026)Gurnee, Ameisen, Kauvar, Tarng, Pearce, Olah, and Batson]{gurnee2026modelsmanipulatemanifoldsgeometry}
Wes Gurnee, Emmanuel Ameisen, Isaac Kauvar, Julius Tarng, Adam Pearce, Chris Olah, and Joshua Batson.
\newblock When models manipulate manifolds: The geometry of a counting task, 2026.
\newblock URL \url{https://arxiv.org/abs/2601.04480}.

\bibitem[Hanna et~al.(2023)Hanna, Liu, and Variengien]{hanna2023does}
Michael Hanna, Ollie Liu, and Alexandre Variengien.
\newblock How does gpt-2 compute greater-than?: Interpreting mathematical abilities in a pre-trained language model.
\newblock \emph{Advances in Neural Information Processing Systems}, 36:\penalty0 76033--76060, 2023.

\bibitem[Hayes et~al.(2026)Hayes, Li, and Andreas]{hayes2026explainingattentionprogramsynthesis}
Amiri Hayes, Belinda~Z Li, and Jacob Andreas.
\newblock Explaining attention with program synthesis, 2026.
\newblock URL \url{https://arxiv.org/abs/2606.19317}.

\bibitem[Hewitt \& Liang(2019)Hewitt and Liang]{hewitt2019controltasks}
John Hewitt and Percy Liang.
\newblock Designing and interpreting probes with control tasks.
\newblock In \emph{Proceedings of the 2019 Conference on Empirical Methods in Natural Language Processing and the 9th International Joint Conference on Natural Language Processing (EMNLP-IJCNLP)}, pp.\  2733--2743, 2019.
\newblock \doi{10.18653/v1/D19-1275}.

\bibitem[Hwang \& Park(2026)Hwang and Park]{hwang2026intrinsic}
Hyeonbin Hwang and Yeachan Park.
\newblock Intrinsic task symmetry drives generalization in algorithmic tasks.
\newblock \emph{arXiv preprint arXiv:2603.01968}, 2026.

\bibitem[Kantamneni \& Tegmark(2025)Kantamneni and Tegmark]{kantamneni2025languagemodelsusetrigonometry}
Subhash Kantamneni and Max Tegmark.
\newblock Language models use trigonometry to do addition, 2025.
\newblock URL \url{https://arxiv.org/abs/2502.00873}.

\bibitem[Karkada et~al.(2026)Karkada, Korchinski, Nava, Wyart, and Bahri]{karkada2026symmetry}
Dhruva Karkada, Daniel~J Korchinski, Andres Nava, Matthieu Wyart, and Yasaman Bahri.
\newblock Symmetry in language statistics shapes the geometry of model representations.
\newblock \emph{arXiv preprint arXiv:2602.15029}, 2026.

\bibitem[Kram{\'a}r et~al.(2024)Kram{\'a}r, Lieberum, Shah, and Nanda]{kramar2024atpstar}
J{\'a}nos Kram{\'a}r, Tom Lieberum, Rohin Shah, and Neel Nanda.
\newblock {AtP*}: An efficient and scalable method for localizing {LLM} behaviour to components.
\newblock \emph{arXiv preprint arXiv:2403.00745}, 2024.

\bibitem[Lee et~al.(2024)Lee, Bai, Pres, Wattenberg, Kummerfeld, and Mihalcea]{lee2024mechanisticunderstandingalignmentalgorithms}
Andrew Lee, Xiaoyan Bai, Itamar Pres, Martin Wattenberg, Jonathan~K. Kummerfeld, and Rada Mihalcea.
\newblock A mechanistic understanding of alignment algorithms: A case study on dpo and toxicity, 2024.
\newblock URL \url{https://arxiv.org/abs/2401.01967}.

\bibitem[Marks \& Tegmark(2024)Marks and Tegmark]{marks2024geometrytruthemergentlinear}
Samuel Marks and Max Tegmark.
\newblock The geometry of truth: Emergent linear structure in large language model representations of true/false datasets, 2024.
\newblock URL \url{https://arxiv.org/abs/2310.06824}.

\bibitem[Marr(2010)]{marr2010vision}
David Marr.
\newblock \emph{Vision: A computational investigation into the human representation and processing of visual information}.
\newblock MIT press, 2010.

\bibitem[McGrath et~al.(2023)McGrath, Rahtz, Kramar, Mikulik, and Legg]{mcgrath2023hydraeffectemergentselfrepair}
Thomas McGrath, Matthew Rahtz, Janos Kramar, Vladimir Mikulik, and Shane Legg.
\newblock The hydra effect: Emergent self-repair in language model computations, 2023.
\newblock URL \url{https://arxiv.org/abs/2307.15771}.

\bibitem[Meng et~al.(2022)Meng, Bau, Andonian, and Belinkov]{meng2022rome}
Kevin Meng, David Bau, Alex Andonian, and Yonatan Belinkov.
\newblock Locating and editing factual associations in {GPT}.
\newblock In \emph{Advances in Neural Information Processing Systems (NeurIPS)}, volume~35, pp.\  17359--17372, 2022.
\newblock arXiv:2202.05262.

\bibitem[Modell et~al.(2025)Modell, Rubin-Delanchy, and Whiteley]{modell2025originsrepresentationmanifoldslarge}
Alexander Modell, Patrick Rubin-Delanchy, and Nick Whiteley.
\newblock The origins of representation manifolds in large language models, 2025.
\newblock URL \url{https://arxiv.org/abs/2505.18235}.

\bibitem[Nanda(2023)]{nanda2023atp}
Neel Nanda.
\newblock Attribution patching: Activation patching at industrial scale.
\newblock \url{https://www.neelnanda.io/mechanistic-interpretability/attribution-patching}, 2023.

\bibitem[Park et~al.(2023)Park, Choe, and Veitch]{park2023linear}
Kiho Park, Yo~Joong Choe, and Victor Veitch.
\newblock The linear representation hypothesis and the geometry of large language models.
\newblock \emph{arXiv preprint arXiv:2311.03658}, 2023.

\bibitem[Syed et~al.(2023)Syed, Rager, and Conmy]{syed2023eap}
Aaquib Syed, Can Rager, and Arthur Conmy.
\newblock Attribution patching outperforms automated circuit discovery.
\newblock In \emph{NeurIPS 2023 Workshop on Attributing Model Behavior at Scale (ATTRIB)}, 2023.
\newblock arXiv:2310.10348.

\bibitem[Tigges et~al.(2023)Tigges, Hollinsworth, Geiger, and Nanda]{tigges2023linearrepresentationssentimentlarge}
Curt Tigges, Oskar~John Hollinsworth, Atticus Geiger, and Neel Nanda.
\newblock Linear representations of sentiment in large language models, 2023.
\newblock URL \url{https://arxiv.org/abs/2310.15154}.

\bibitem[Vig et~al.(2020)Vig, Gehrmann, Belinkov, Qian, Nevo, Singer, and Shieber]{vig2020causal}
Jesse Vig, Sebastian Gehrmann, Yonatan Belinkov, Sharon Qian, Daniel Nevo, Yaron Singer, and Stuart Shieber.
\newblock Investigating gender bias in language models using causal mediation analysis.
\newblock In \emph{Advances in Neural Information Processing Systems (NeurIPS)}, volume~33, pp.\  12388--12401, 2020.

\bibitem[Voynov \& Babenko(2020)Voynov and Babenko]{voynov2020unsuperviseddiscoveryinterpretabledirections}
Andrey Voynov and Artem Babenko.
\newblock Unsupervised discovery of interpretable directions in the gan latent space, 2020.
\newblock URL \url{https://arxiv.org/abs/2002.03754}.

\bibitem[Wang et~al.(2023)Wang, Variengien, Conmy, Shlegeris, and Steinhardt]{wang2023ioi}
Kevin Wang, Alexandre Variengien, Arthur Conmy, Buck Shlegeris, and Jacob Steinhardt.
\newblock Interpretability in the wild: A circuit for indirect object identification in {GPT-2} small.
\newblock In \emph{International Conference on Learning Representations (ICLR)}, 2023.
\newblock arXiv:2211.00593.

\bibitem[Wu et~al.(2024)Wu, Geiger, Icard, Potts, and Goodman]{wu2024interpretabilityscaleidentifyingcausal}
Zhengxuan Wu, Atticus Geiger, Thomas Icard, Christopher Potts, and Noah~D. Goodman.
\newblock Interpretability at scale: Identifying causal mechanisms in alpaca, 2024.
\newblock URL \url{https://arxiv.org/abs/2305.08809}.

\bibitem[Wurgaft et~al.(2026)Wurgaft, Rager, Kowal, Shyam, Feucht, Bhalla, Haklay, Bigelow, Sarfati, McGrath, Lewis, Merullo, Goodman, Fel, Geiger, and Lubana]{wurgaft2026manifoldsteeringrevealsshared}
Daniel Wurgaft, Can Rager, Matthew Kowal, Vasudev Shyam, Sheridan Feucht, Usha Bhalla, Tal Haklay, Eric Bigelow, Raphael Sarfati, Thomas McGrath, Owen Lewis, Jack Merullo, Noah Goodman, Thomas Fel, Atticus Geiger, and Ekdeep~Singh Lubana.
\newblock Manifold steering reveals the shared geometry of neural network representation and behavior, 2026.
\newblock URL \url{https://arxiv.org/abs/2605.05115}.

\bibitem[Yang et~al.(2024)Yang, Yang, Zhang, Hui, Zheng, Yu, Li, Liu, Huang, Wei, et~al.]{qwen25}
An~Yang, Baosong Yang, Beichen Zhang, Binyuan Hui, Bo~Zheng, Bowen Yu, Chengyuan Li, Dayiheng Liu, Fei Huang, Haoran Wei, et~al.
\newblock Qwen2.5 technical report.
\newblock \emph{arXiv preprint arXiv:2412.15115}, 2024.

\bibitem[Yuchi et~al.(2026)Yuchi, Du, and Eisner]{yuchi2026llmsknownumberssay}
Fengting Yuchi, Li~Du, and Jason Eisner.
\newblock Llms know more about numbers than they can say, 2026.
\newblock URL \url{https://arxiv.org/abs/2602.07812}.

\bibitem[Zhu et~al.(2024)Zhu, Zhang, and Wang]{zhu2024languagemodelsrepresentbeliefs}
Wentao Zhu, Zhining Zhang, and Yizhou Wang.
\newblock Language models represent beliefs of self and others, 2024.
\newblock URL \url{https://arxiv.org/abs/2402.18496}.

\end{thebibliography}
\bibliographystyle{iclr2027_conference}

\newpage
\appendix




\section{Experimental Setup}
\label{app:setup}

\subsection{Model, task and prompts}
\label{app:setup-model}
\label{app:setup-prompt}
\label{app:setup-tokens}
\label{app:setup-regimes}

We use Qwen2.5-7B-Instruct \citep{qwen25} ($28$ layers, $d_{\text{model}} = 3584$, $28$ heads of
dimension $128$, SwiGLU MLPs of width $d_{\text{ff}} = 18944$) in \texttt{float16} with all
parameters frozen, on a single NVIDIA A100 40GB GPU. The task $\max(y_1,\dots,y_K)$ is posed as a raw
completion (no chat template) with a one-shot exemplar that fixes the answer format:
\begin{quote}\ttfamily\small
Answer in the following format with a single answer. The maximum of 12 and 4 is 12. The maximum of \{$y_1$\} and \{$y_2$\} is
\end{quote}
\noindent for $K=2$ in both digit regimes, with the exemplar ``\texttt{The maximum of 12, 437 and 5 is 437.}'' and a
question ``\texttt{The maximum of \{$y_1$\}, \{$y_2$\} and \{$y_3$\} is}'' for $K \ge 3$; each
prompt ends in a single space.

Qwen2.5 tokenizes numbers digit by digit, so an operand's first token is its leading digit. We read
every residual-stream quantity at a number's last token, its \emph{position}, where the whole number
is first visible under causal attention; all operands are first visible at the position of $y_2$
($K=2$) or $y_3$ ($K=3$). Positions are found from the operands' character spans after the last
occurrence of \texttt{"The maximum of "} via the tokenizer's offset mapping. Within an experiment
all operands have the same number of digits, so every prompt has the same token layout ($T=36$ with
$y_1, y_2$ at tokens $29, 33$ for two 2-digit operands; $T=38$ with tokens $30, 35$ for two 3-digit
operands; $T=46$ with tokens $35, 39, 43$ for three 2-digit operands). All sampled operands have
pairwise-distinct leading digits, so that each readout token identifies one operand. 

Two-digit operands lie in $[10, 100)$ with a minimum gap $g = 10$ between the values of a sampled
tuple, and three-digit operands in $[100, 1000)$ with $g = 100$; the three-digit regime is used only
to replicate the number-representation analysis.

\subsection{Counterfactual pairs}
\label{app:setup-cf2}
\label{app:setup-readout}
\label{app:setup-cf3}

For $K=2$ we sample triples $(a, b, r)$ which satisfy $a > b + g > r + 2g$ ($g$ is the gap). The clean prompt (e.g., $(a, b)$) has $a$ at the winning operand and
$b$ at the other (answer is $a$), and the corrupted prompt replaces $a$ in place by $r$ (so corrupted prompt has ($r, b$)) (answer $b$), so
that only which operand wins changes. The same triples are used in both arrangements (only the order of values is changed), $y_1 > y_2$ and
$y_2 > y_1$. Patching a clean state into the corrupted run makes the model name $r$ (smaller value in its input pair) rather than $a$ (Section~\ref{app:res-behaviour}): the patch
changes which operand the model treats as the maximum, not the value it outputs. We therefore score
interventions on $t(r)$ against $t(b)$, where $t(\cdot)$ is a number's first token; unlike $t(a)$,
$t(r)$ cannot be raised by a patch that merely re-inserts the clean value.

For $K=3$ we sample quadruples $a > b > c > r$ with consecutive gaps larger than $g$, place $a, b, c$
in a chosen value order in the clean prompt, and replace $a$ in place by $r$ in the corrupted one, so
that the corrupted winner is always the clean runner-up. The \emph{head groups} $y_1 > y_2 > y_3$
(patched at the position of $y_2$), $y_1 > y_3 > y_2$ and $y_2 > y_3 > y_1$ (both patched at the
position of $y_3$) vary the runner-up's position and the maximum's position one at a time and are
used to identify heads. All other analyses use three \emph{perturbation cases}, patched at the
position of $y_3$ and named by the number the corruption changes: case $y_1$ ($y_1 > y_3 > y_2$),
case $y_2$ ($y_2 > y_3 > y_1$) and case $y_3$ ($y_3 > y_2 > y_1$).

\subsection{Populations and splits}
\label{app:setup-splits}
\label{app:setup-behaviour}

Intervention scores are computed on $400$ evaluation examples, and directions and neuron rankings
are fitted on $128$ disjoint fitting examples, both taken from one deduplicated pool drawn by a
deterministic sampler, so that every analysis uses the same examples. For $K = 3$ we draw $1{,}400$
quadruples, keep the $1{,}341$ whose clean and corrupted prompts are solved (next-token argmax $t(a)$
and $t(b)$) in all four value orders, following \citet{meng2022rome}, and use the first $528$. For
$K = 2$ no filter is applied: in the two-digit regime the model answers every ordered pair of distinct
two-digit numbers correctly, while the three-digit populations include one example per arrangement
with a non-positive clean-minus-corrupted gap. Geometry and probes use separate clouds of clean
prompts, with the two-number clouds containing each unordered pair in both orders.

Behavioural accuracy is measured by greedy decoding of one token more than the operands' digit count,
comparing the first integer of the output with the true maximum, on all $8{,}010$ ordered pairs of
distinct two-digit numbers and on $4{,}000$ three-digit pairs drawn with replacement. The
operand-count sweep uses $200$ prompts of distinct two-digit operands per length in the three-operand
prompt format, with Wilson score intervals \citep{brown2001interval}. Table~\ref{app:setup-hparams}
lists all sample sizes and hyperparameters.

\begin{table}[!ht]
\centering
\small
\caption{Sample sizes and hyperparameters.}
\label{app:setup-hparams}
\begin{tabular}{llc}
\toprule
& quantity & value \\
\midrule
\multirow{4}{*}{\emph{populations}}
& evaluation / fitting examples                  & $400$ / $128$ \\
& clouds (L13 two-number / L14 / three-number)   & $2000$ / $3000$ / $1500$ \\
& operand counts $k$ in the accuracy sweeps, prompts per $k$ & $2, 3, 4, 5, 10, 20$; $200$ \\
& base random seed                               & $52$ \\
\midrule
\multirow{3}{*}{\emph{learned directions}}
& Adam steps, learning rate                      & $100$, $0.05$ \\
& minibatch (gradients accumulated)              & $16$--$32$ \\
& initializations in the stability check ($\bm{u}$) & $2$ \\
\midrule
\multirow{3}{*}{\emph{attribution}}
& minibatch, half-precision loss scale           & $16$, $100$ \\
& candidate neurons (MLP14 $\cup$ MLP15)         & $37{,}888$ \\
& freeze-sweep depths $k$                        & $1, 3, 10, 30, 10^2, 3\!\cdot\!10^2, 10^3, 3\!\cdot\!10^3$ \\
\midrule
\multirow{3}{*}{\emph{sweeps and fields}}
& dose-response grid values, window              & $15, 25, \dots, 95$; $\pm 2$ \\
& prompts per dose-response cell ($K{=}2$)       & $64$ \\
& receptive-field grid                           & stride $4$ over $y \in [11, 99]$ \\
\midrule
\multirow{2}{*}{\emph{probes}}
& ridge penalty grid (leave-one-out)             & $13$ log-spaced values in $[10^{-2}, 10^{4}]$ \\
& held-out fraction                              & $0.2$ \\
\bottomrule
\end{tabular}
\end{table}


\section{Methods}
\label{app:methods}

\subsection{Interventions and sites}
\label{app:methods-ii}
\label{app:methods-sites}

Causal measurements are interchange interventions \citep{geiger2021causal, geiger2023das}: we run the
corrupted prompt and replace a component of the activation at one site by its value on the clean
run. With $\bm{x}$ the corrupted activation, $\bm{x}^{\text{clean}}$ the clean one and
$P \in \mathbb{R}^{d \times k}$ an orthonormal basis,
\begin{equation}
\bm{x}' \;=\; \bm{x} \;+\; P P^{\top}\!\left(\bm{x}^{\text{clean}} - \bm{x}\right),
\label{eq:interchange}
\end{equation}
computed in single precision. $P = I$ is the full-rank restoration of \citet{meng2022rome}. Each
rank-restricted patch is shown next to a full-rank patch at the same site as a reference rather than
an upper bound: rank-restricted patches can exceed it.

The sites are: (i)~the residual stream leaving a decoder block, at one position; (ii)~one attention
head's contribution $\bm{z}_h (W_O^h)^{\top}$, obtained from the head's $128$-dimensional slice of
the output-projection input, where a direction $P$ is patched by reading the coefficient
$(\bm{z}_h^{\text{clean}} - \bm{z}_h)(W_O^h)^{\top}P$ from the head's contribution and adding it
along $P$ to the residual stream ($P$ is not constrained to the head's output space; for $\bm{v}_1$ in
L14.H14, $81\%$ of its norm lies inside it); (iii)~the pre-MLP residual
$\bm{x}_{\ell-1} + \mathrm{Attn}_\ell(\bm{x}_{\ell-1})$, patched through the attention output, which
gives the same result as an end-of-layer patch at full rank; and (iv)~MLP neurons, the coordinates of
the post-SwiGLU vector that feeds the down-projection.

\subsection{Metrics}
\label{app:methods-metrics}

With $\mathrm{PLD} = \mathrm{logit}[t(r)] - \mathrm{logit}[t(b)]$ computed at the last token, using
the same token pair on the clean, corrupted and patched runs, \emph{position recovery} is
\begin{equation}
\mathrm{PR} \;=\; \frac{\mathrm{PLD}_{\text{patched}} - \mathrm{PLD}_{\text{corr}}}
                            {\mathrm{PLD}_{\text{clean}} - \mathrm{PLD}_{\text{corr}}},
\end{equation}
the normalized restoration of \citet{meng2022rome} applied to the logit difference of
\citet{wang2023ioi}, computed per example and averaged. The numerator is the indirect effect of the
patched component on the PLD \citep{vig2020causal}, and the denominator is a per-example logit scale
of $10$--$13$ logits on average, which makes values comparable across sites and cases but is not
meant to make a full effect equal $1$: since $t(r)$ is not an answer on the clean prompt,
$\mathrm{PLD}_{\text{clean}}$ can be negative. Where IIA is near zero, as in three-number cases $y_1$ and $y_2$, PR measures how far
the PLD moves rather than a change of answer. PR need not lie in $[0,1]$.

\emph{Interchange intervention accuracy} \citep{geiger2021causal} is
$\mathrm{IIA} = \mathbb{I}\left[\arg\max_{v \in \mathcal{V}} \mathrm{logit}[v] = t(r)\right]$ over the
full vocabulary at the last token, which scores the first generated token, the leading digit of $r$,
so it also counts the mixed answers produced by a full-rank patch at the last token of $y_2$
(Section~\ref{app:res-behaviour}).
For the ablations we use \emph{full-number accuracy}: the answer is decoded greedily for one token
more than the operands' digit count and counted correct when its leading characters match the true
maximum.

\subsection{Directions}
\label{app:methods-das}
\label{app:methods-frames}
\label{app:methods-baselines}

Learned directions are rank-$k$ subspaces found by distributed alignment search for a single causal
variable \citep{geiger2023das}: the basis $Q = \mathrm{qr}(R)$ (using QR factorization) of a free matrix $R$ is inserted into
Equation~\ref{eq:interchange} at the chosen site and trained on the fitting examples with a
cross-entropy toward $t(r)$ restricted to $\{t(b), t(r)\}$, with gradients accumulated so that each
step uses the full-batch mean. With $16$ fitting examples a fit reached a training IIA of $1.000$
against a held-out $0.688$, hence the $128$ used here. Retraining $\bm{u}$ from a second
initialization gives $|\cos| = 0.99$ and the same held-out IIA ($0.94$) for two-digit operands, and
$|\cos| = 0.82$ with IIAs $0.86$ and $0.80$ for three-digit operands; other directions use a single
initialization. Where a site may carry one variable shared across cases, a subspace fitted on each
case is evaluated on every case. Directions are signed so that their component rises with the number
they carry. Table~\ref{tab:app-directions} lists the directions.

\begin{table}[!ht]
\centering
\small
\caption{Directions, where they live and how they are obtained. DAS directions are rank one and
fitted in the listed case; PC denotes the top principal component(s) of the corresponding clean
cloud.}
\label{tab:app-directions}
\begin{tabular}{lllll}
\toprule
$K$ & direction & position & site & obtained by \\
\midrule
$2$ & $\bm{u}$ & $y_1$ & L13 residual & DAS, $y_1$ perturbed \\
$2$ & $\bm{v}_1$ & $y_2$ & output of L14.H14 & DAS, $y_1$ perturbed \\
$2$ & $\bm{v}_2$ & $y_2$ & L13 residual & PC \\
$2$ & comparison flag & $y_2$ & L15 residual & DAS, each case \\
$3$ & $\bm{u}_1$, $\bm{u}_2$ & $y_1$, $y_2$ & L13 residual & DAS, cases $y_1$, $y_2$ \\
$3$ & $\bm{v}_1$, $\bm{v}_2$ & $y_3$ & outputs of L14.H14, L14.H18 & DAS, cases $y_1$, $y_2$ \\
$3$ & $\bm{v}_3$ & $y_3$ & L13 residual & PC \\
$3$ & $\bm{w}_1$, $\bm{w}_2$ & $y_2$ & L13 residual & DAS, cases $y_1$, $y_2$ ($\bm{w}_2 = \bm{u}_2$) \\
$3$ & comparison flags & $y_2$, $y_3$ & L15 residual & DAS, each case \\
$3$ & answer subspace & last token & L20 residual & top two PCs \\
\bottomrule
\end{tabular}
\end{table}

$\bm{v}_2$ is a principal component because a direction learned at that site partly encodes the
comparison outcome instead of $y_2$ (Section~\ref{app:res-numbers}), and heads L14.H14 and L14.H18
come from our earlier localization of the three-number task. Multi-dimensional patches use the span
of the $\bm{v}_j$, orthogonalized by Gram--Schmidt with $\bm{v}_1$ first; the directions are nearly
orthogonal to begin with ($\cos(\bm{v}_1, \bm{v}_2) = -0.08$ for two numbers), and because they are
obtained independently, the span's score is a lower bound for a rank-$k$ patch at that site.

Principal components are fitted on mean-centred, unscaled activations \citep{gurnee2023space}.
Linear decodability is measured with ridge regression \citep{alain2016probes, belinkov2022survey}
on the logarithm of each number, and the comparison $y_1 > y_2$ with a ridge classifier, reporting
held-out $R^2$ or accuracy; probes for operands that a position cannot depend on serve as floors, in
the spirit of the control tasks of \citet{hewitt2019controltasks}. Log-linear trends are summarized by
the least-squares fit $p \log y + q$.

\subsection{Localization}
\label{app:methods-localization}
\label{app:methods-readout}

The causal trace \citep{meng2022rome} restores the clean residual stream leaving one layer at one
token, for every layer and every token from the first token of $y_1$ to the last token of the prompt,
and reports position recovery averaged within each arrangement. Heads are then interchanged at site
(ii) of layer 14 \citep{wang2023ioi, goldowskydill2023pathpatching}, each together with a restoration
of the layer-13 residual at the same position (in full for three numbers, along $\bm{v}_2$ for two
numbers), since a single head written into a corrupted residual stream has almost no effect; the
co-patch alone is shown as the baseline, and for two numbers the sweep is also shown without it. At
the last token, full-rank patches of the residual leaving layers $18$ to $22$ over the head-group
orders locate the read-out at layer $20$, where mean position recovery increases most ($0.09$ to
$0.55$). At that layer we patch each head's contribution without a co-patch, with the whole attention
sublayer as a reference, and compare a patch of the top two principal components of the residual
with a full-rank patch.

\subsection{Neurons}
\label{app:methods-attribution}
\label{app:methods-freeze}

We rank the neurons of MLP14 and MLP15 by attribution patching
\citep{nanda2023atp, syed2023eap, kramar2024atpstar}, the first-order estimate of the effect on the
PLD of freezing neuron $i$ while the shared representation is patched,
$s_i \approx (a_i^{\text{corr}} - a_i^{\text{patched}})\,\partial\mathrm{PLD}/\partial a_i$, with the
gradient taken at the patched state through a zero-valued differentiable input at the
down-projection. Rankings are computed on the fitting examples and verified by freezing the top $k$
neurons on the evaluation examples, against two random draws of $k$ neurons. The shared set is the
intersection of the top-$20$ rankings of the two two-number cases, $12$ neurons (six in each MLP),
which the three-number analysis reuses unchanged.

To freeze a sub-block or a set of neurons, we pin its output (for neurons, their post-SwiGLU values)
at the target position to its value on the unpatched corrupted run, the severed-path variant of causal
tracing \citep{meng2022rome, vig2020causal}, for the attention and MLP sub-blocks of layers $14$ to
$16$ individually and for MLP14 and MLP15 jointly. For injection we write only the clean outputs of
the selected MLPs into the corrupted run at the target position. For ablation we set the shared
neurons' post-SwiGLU activations to zero, at every position or only at the last token of the final
operand, throughout greedy decoding, and compare with two random sets of $12$ neurons zeroed at every
position, scoring full-number accuracy.

\subsection{Dose-response sweeps, receptive fields and connectivity}
\label{app:methods-sweep}
\label{app:methods-rf}
\label{app:methods-connect}
\label{app:methods-nulls}

For the dose-response sweep \citep{geiger2021causal, chan2022scrubbing} we set the two coordinates of
the $(\bm{v}_1, \bm{v}_2)$ frame in the pre-MLP layer-14 residual at the position of $y_2$ to the mean
coordinate of a chosen number (over a $\pm 2$ window in the cloud), on a $9 \times 9$ grid with one
value per leading digit, and score the fraction of argmax answers that name the first operand among
those naming either operand, masking the diagonal. The pre-MLP site is used because a per-number mean
explains $R^2 = 0.87$ of the $\bm{v}_1$ coordinate there, against $0.35$ at the end of the layer.

A neuron's receptive field is its post-SwiGLU activation at the target position over a grid of
prompts, one forward pass per cell and without the distinct-leading-digit constraint. Three-number
fields are shown as pairwise marginals, each averaged over the third number and divided by its own
peak, and an operand that a position cannot depend on is held fixed ($y_3 = 55$ at the position of
$y_2$).

The causal edge from an MLP14 neuron to an MLP15 neuron is the shift in the MLP15 neuron's activation
when only the MLP14 neuron is set to its clean value, in units of the MLP15 neuron's standard
deviation over a $1{,}500$-prompt two-number cloud. The virtual weight \citep{elhage2021framework}
through the gate projection is the cosine between the upstream write column and the downstream gate
read row with the RMSNorm scale folded in, and the alignment of a write direction $\bm{w}_j$ with a
learned subspace $Q$ is $\rho_j = \lVert Q^{\top}\bm{w}_j\rVert / \lVert \bm{w}_j \rVert$, read against
the distribution over all $18{,}944$ neurons of the layer, since write directions are not isotropic
(the analytic level for a random unit vector and a $k$-dimensional subspace is about
$\sqrt{k/d_{\text{model}}}$). As a specificity control, each three-number direction is re-evaluated,
without refitting, in the cases where the number it carries does not change, where it should have no
effect.

\newpage
\providecommand{\FloatBarrier}{\clearpage}

\section{Further Experimental Results}
\label{app:results}

The figures below follow the order of the main-text algorithm. Unless a caption says otherwise,
scores are means over the $400$ held-out counterfactuals of Section~\ref{app:setup-splits}, error
bars are one standard error, and PR (position recovery) and IIA (interchange intervention
accuracy) are as defined in Section~\ref{app:methods-metrics}.

\subsection{Behaviour and what an intervention changes}
\label{app:res-behaviour}

We first report task accuracy and the answers the model generates under two interventions.

\begin{figure}[!ht]
  \centering
  \includegraphics[width=0.5\linewidth]{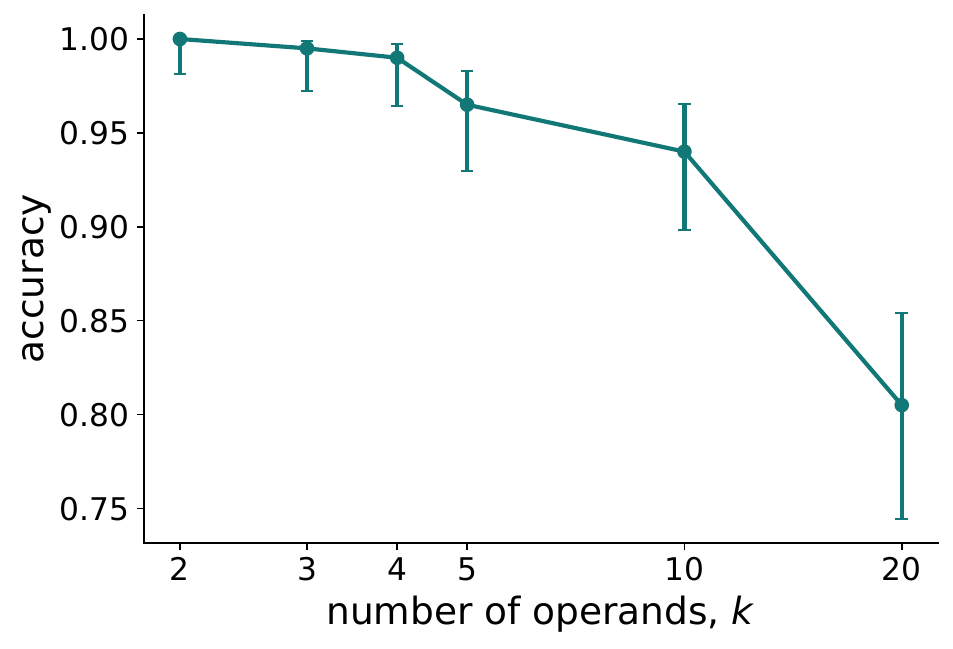}
  \caption{Accuracy on $\max(y_1,\dots,y_k)$ against the number of operands $k$, under greedy decoding, for $200$ prompts of distinct two-digit operands per length. Error bars are Wilson score intervals.}
  \label{fig:app-behaviour}
\end{figure}

\begin{figure}[!ht]
  \centering
  \includegraphics[width=0.75\linewidth]{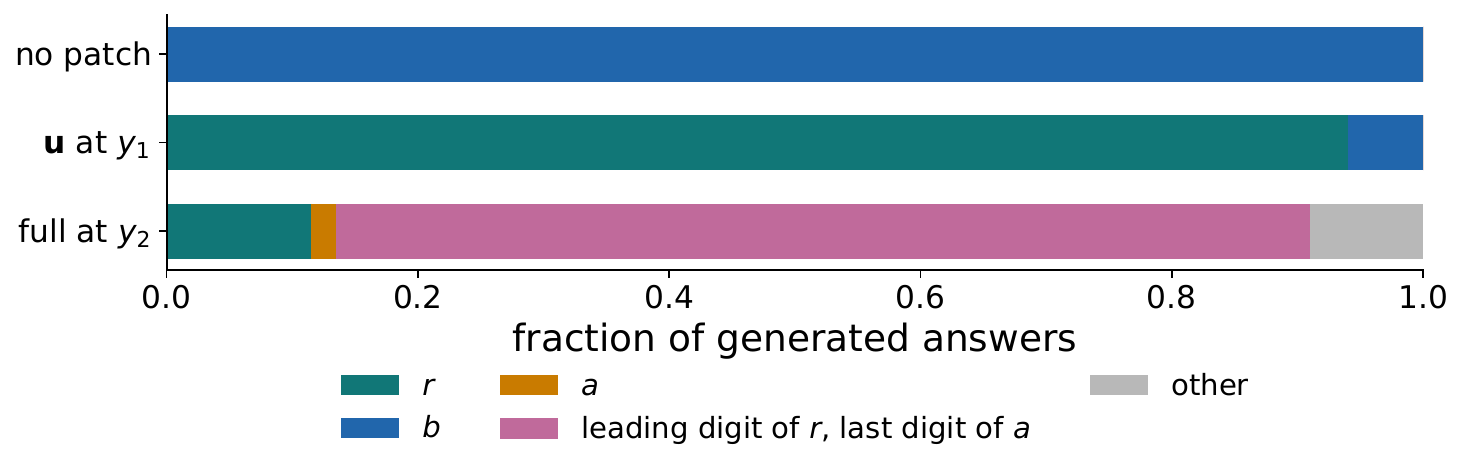}
  \caption{Answers generated by greedy decoding on the $400$ held-out counterfactuals, as fractions per category: $r$, $b$, $a$, the leading digit of $r$ followed by the last digit of $a$, and other. Rows: the corrupted prompt with no patch and with the full layer-13 residual patched at the last token of $y_2$ (both with $y_2$ perturbed), and with $\mathbf{u}$ patched at the position of $y_1$ ($y_1$ perturbed). An answer equal to $r$ is counted as $r$.}
  \label{fig:app-patch-outcomes}
\end{figure}

\begin{table}[!ht]
  \centering
  \small
  \caption{Examples of generated answers on held-out counterfactuals: the clean and corrupted operands, their maxima, and the answer generated under each intervention.}
  \label{tab:app-patch-examples}
  \begin{tabular}{lccccc}
\toprule
intervention & clean $(y_1, y_2)$ & max & corrupted $(y_1, y_2)$ & max & patched answer \\
\midrule
$\mathbf{u}$ at $y_1$ & (99, 61) & 99 & (12, 61) & 61 & 12 \\
$\mathbf{u}$ at $y_1$ & (70, 55) & 70 & (33, 55) & 55 & 33 \\
full at $y_2$ & (61, 99) & 99 & (61, 12) & 61 & 12 \\
full at $y_2$ & (56, 97) & 97 & (56, 27) & 56 & 27 \\
full at $y_2$ & (55, 70) & 70 & (55, 33) & 55 & 30 \\
full at $y_2$ & (71, 89) & 89 & (71, 52) & 71 & 59 \\
\bottomrule
\end{tabular}

\end{table}

\FloatBarrier
\subsection{Where the computation happens}
\label{app:res-localization}

Causal traces and linear probes over every layer and question token
(Section~\ref{app:methods-localization}) locate the sites analysed in the rest of the appendix.

\begin{figure}[!ht]
  \centering
  \includegraphics[width=0.85\linewidth]{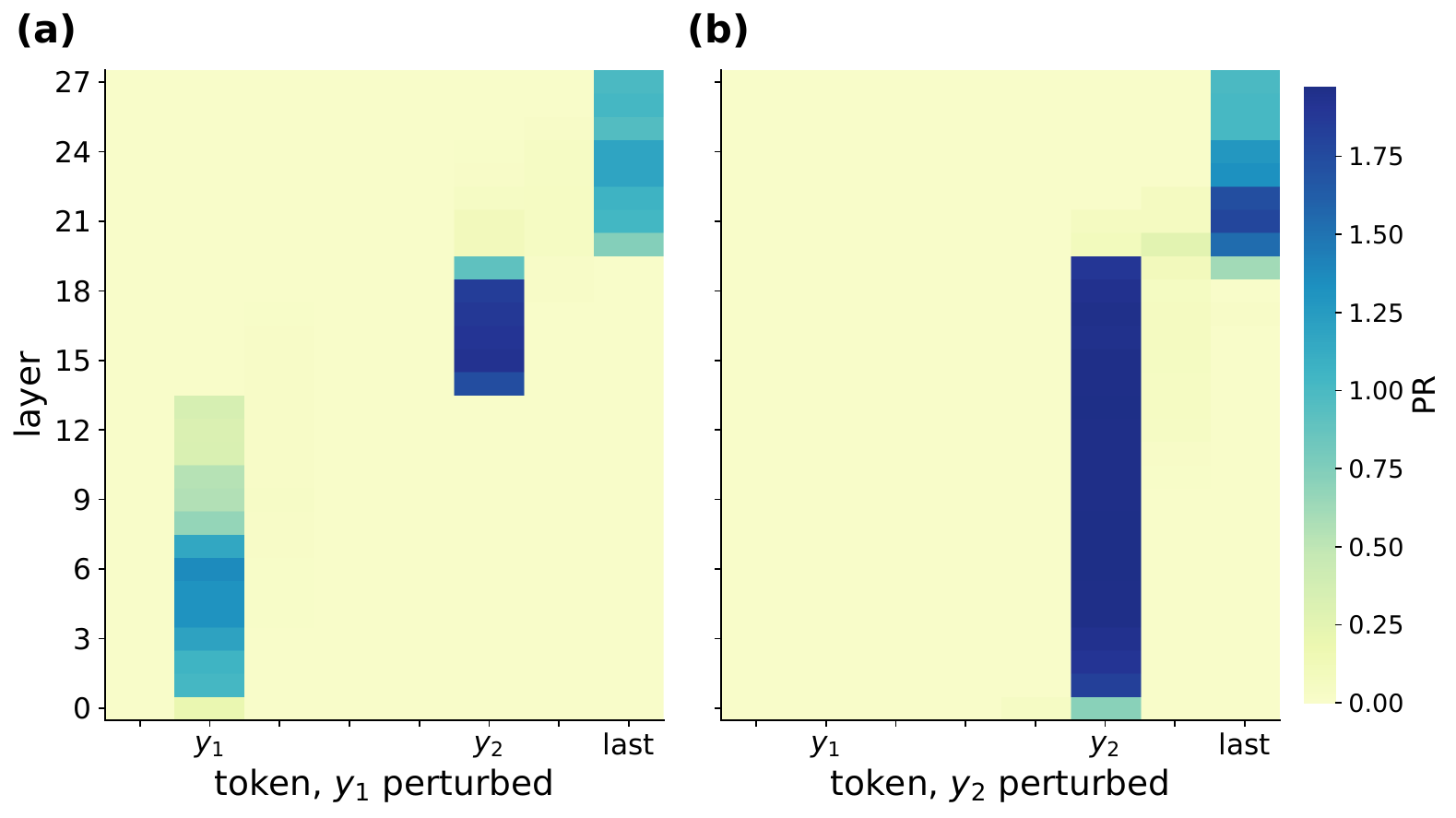}
  \caption{Position recovery when the clean residual stream leaving one layer (rows) is restored at one token (columns), from the first token of $y_1$ to the last token of the prompt, with (a)~$y_1$ and (b)~$y_2$ perturbed. Each operand spans two tokens and the tick marks its last one. The panels share a colour scale.}
  \label{fig:app-trace-k2}
\end{figure}

\begin{figure}
    \centering
    \includegraphics[width=\linewidth]{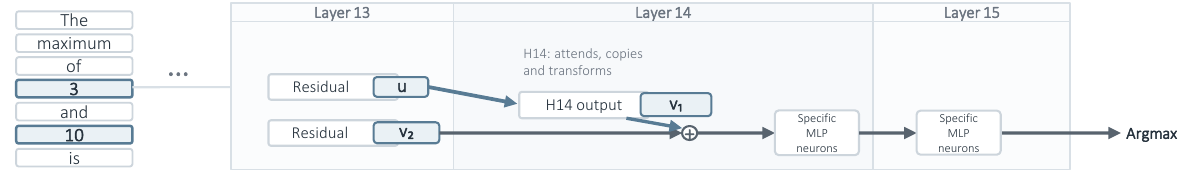}
    \caption{Illustration of the key components implementing algorithm underlying the pairwise number comparisons algorithm Alg. \ref{alg:pairwise-compare}}
    \label{fig:alg-flowchart}
\end{figure}

\begin{figure}[!ht]
  \centering
  \includegraphics[width=\linewidth]{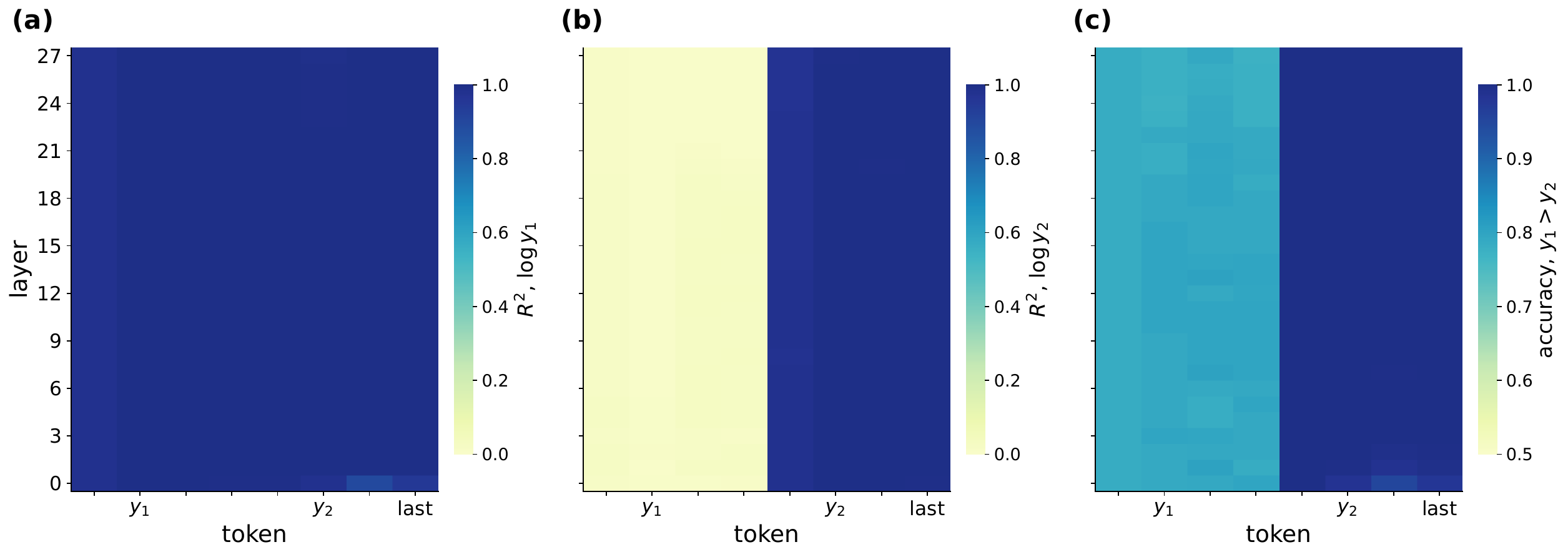}
  \caption{Ridge probes fitted at every layer (rows) and token (columns) on $2{,}000$ clean two-number prompts and scored on a $20\%$ held-out split. (a,b)~Held-out $R^2$ for $\log y_1$ and $\log y_2$. (c)~Held-out accuracy of a ridge classifier for $y_1>y_2$. The axes are those of Figure~\ref{fig:app-trace-k2}.}
  \label{fig:app-probes-k2}
\end{figure}

\begin{figure}[!ht]
  \centering
  \includegraphics[width=\linewidth]{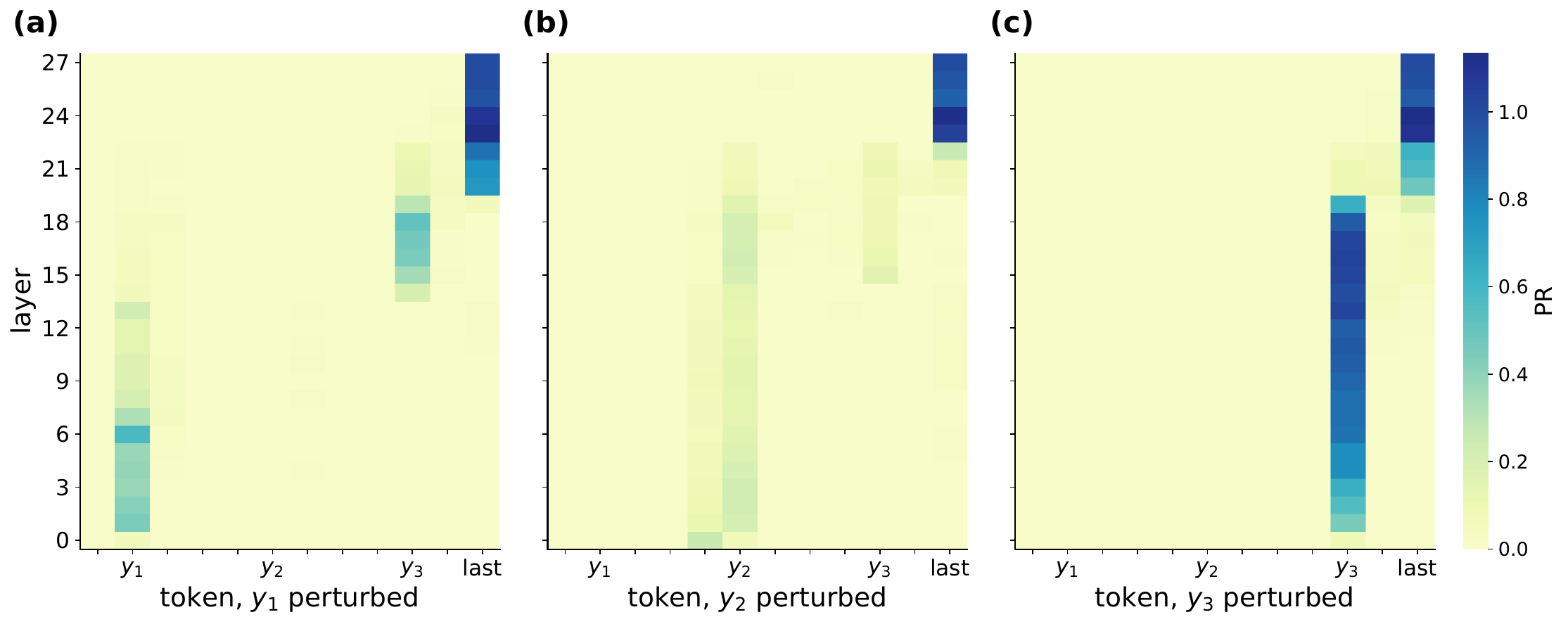}
  \caption{The causal trace of Figure~\ref{fig:app-trace-k2} for three numbers, with (a)~$y_1$, (b)~$y_2$ and (c)~$y_3$ perturbed, on the $400$ held-out quadruples. The panels share a colour scale.}
  \label{fig:app-trace-k3}
\end{figure}

\FloatBarrier
\subsection{Individual number representations}
\label{app:res-numbers}

For the direction $\mathbf{u}$ at the position of $y_1$, we show its place in the layer-13 geometry,
a comparison with the top principal component, the same analysis at earlier layers and with
three-digit operands. We then show the learned alternative to $\mathbf{v}_2$ and the output of the
transport head L14.H14.

\begin{figure}[!ht]
  \centering
  \includegraphics[width=0.85\linewidth]{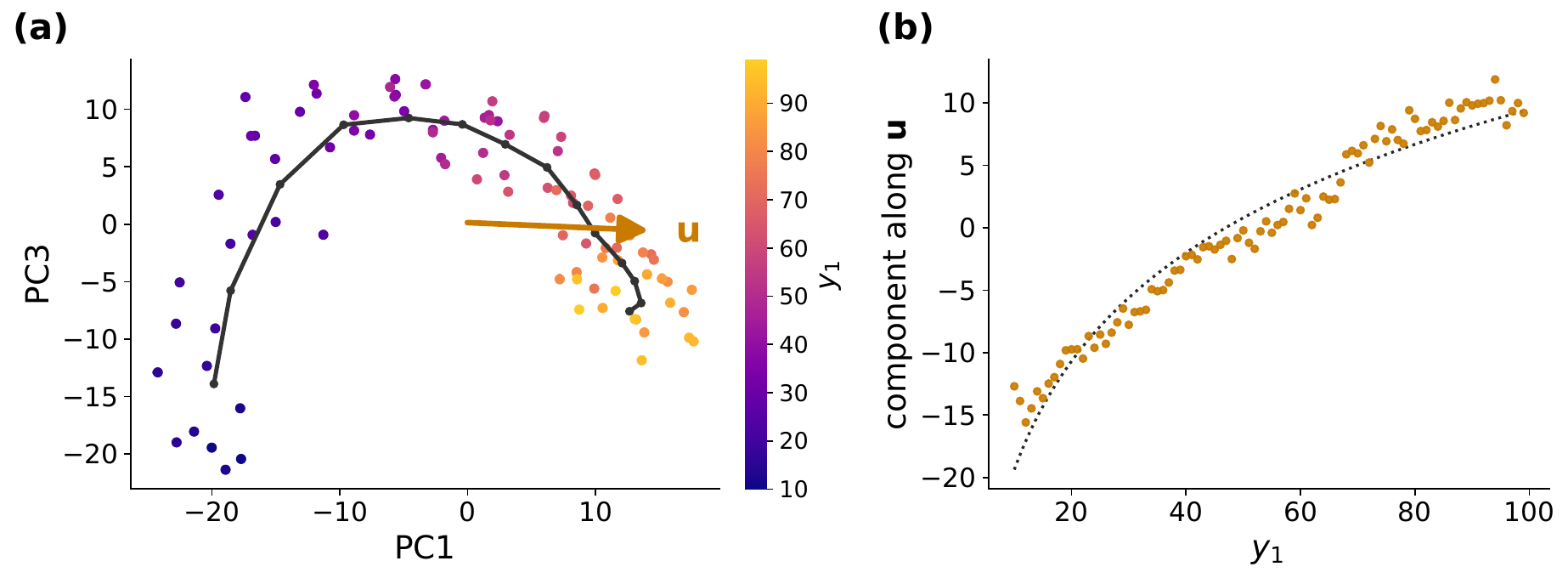}
  \caption{The layer-13 residual at the position of $y_1$, one point per value of $y_1$. (a)~Projection on the first and third principal components, coloured by $y_1$, with the smoothed mean position along $y_1$ (line) and the direction of $\mathbf{u}$'s projection onto the plane (arrow). (b)~Component along $\mathbf{u}$ against $y_1$, with the least-squares fit $p\log y_1+q$ (dotted).}
  \label{fig:app-u-manifold}
\end{figure}

\begin{figure}[!ht]
  \centering
  \includegraphics[width=0.8\linewidth]{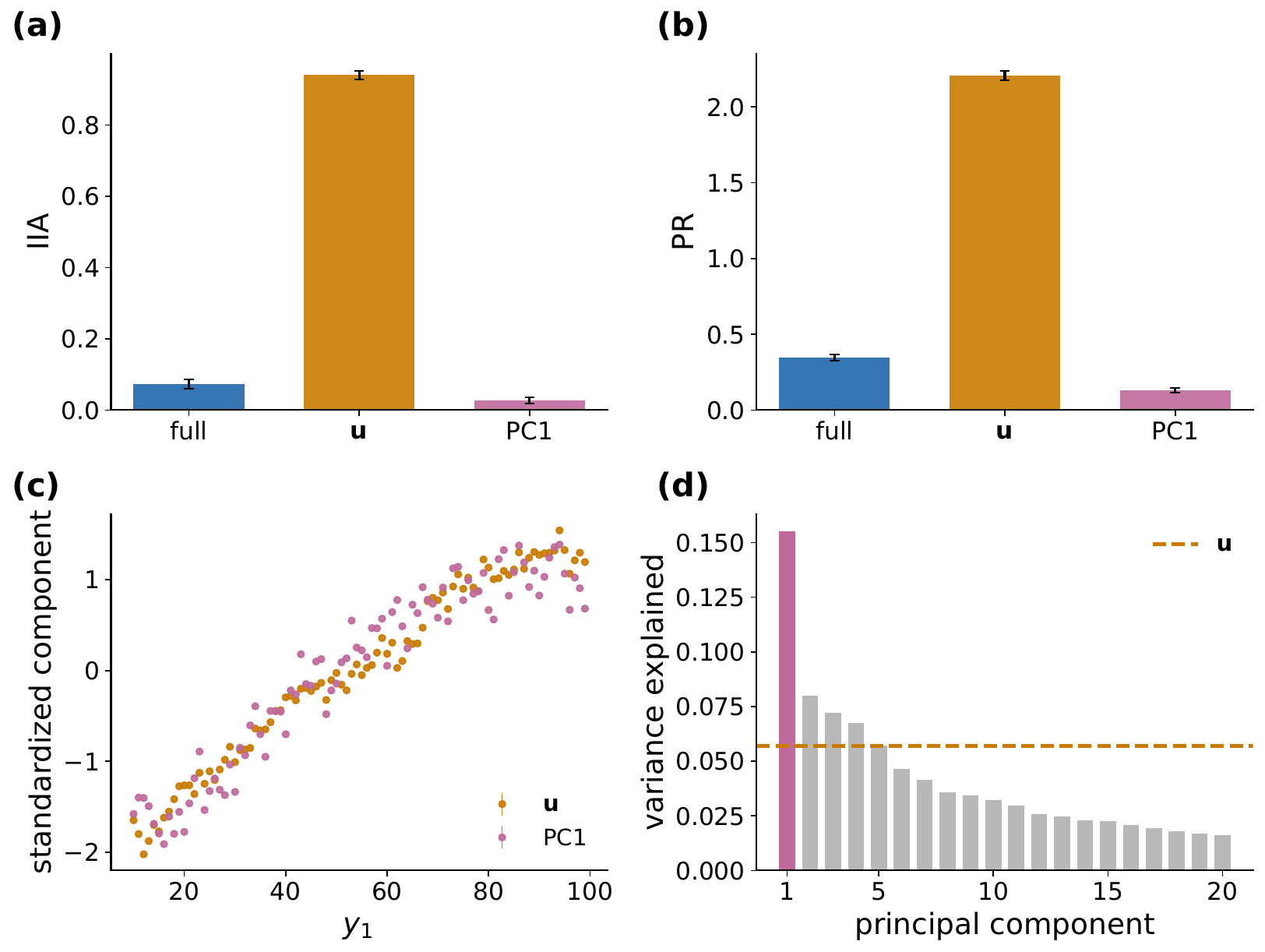}
  \caption{The layer-13 residual at the position of $y_1$, with $y_1$ perturbed. (a)~IIA and (b)~position recovery of a full-rank patch, a rank-one patch of $\mathbf{u}$ and a rank-one patch of the top principal component (PC1). (c)~Standardized components along $\mathbf{u}$ and PC1 against $y_1$, as per-value means. (d)~Variance explained by each of the first $20$ principal components (bars) and by $\mathbf{u}$ (dashed line).}
  \label{fig:app-u-vs-pc1}
\end{figure}

\begin{figure}[!ht]
  \centering
  \includegraphics[width=\linewidth]{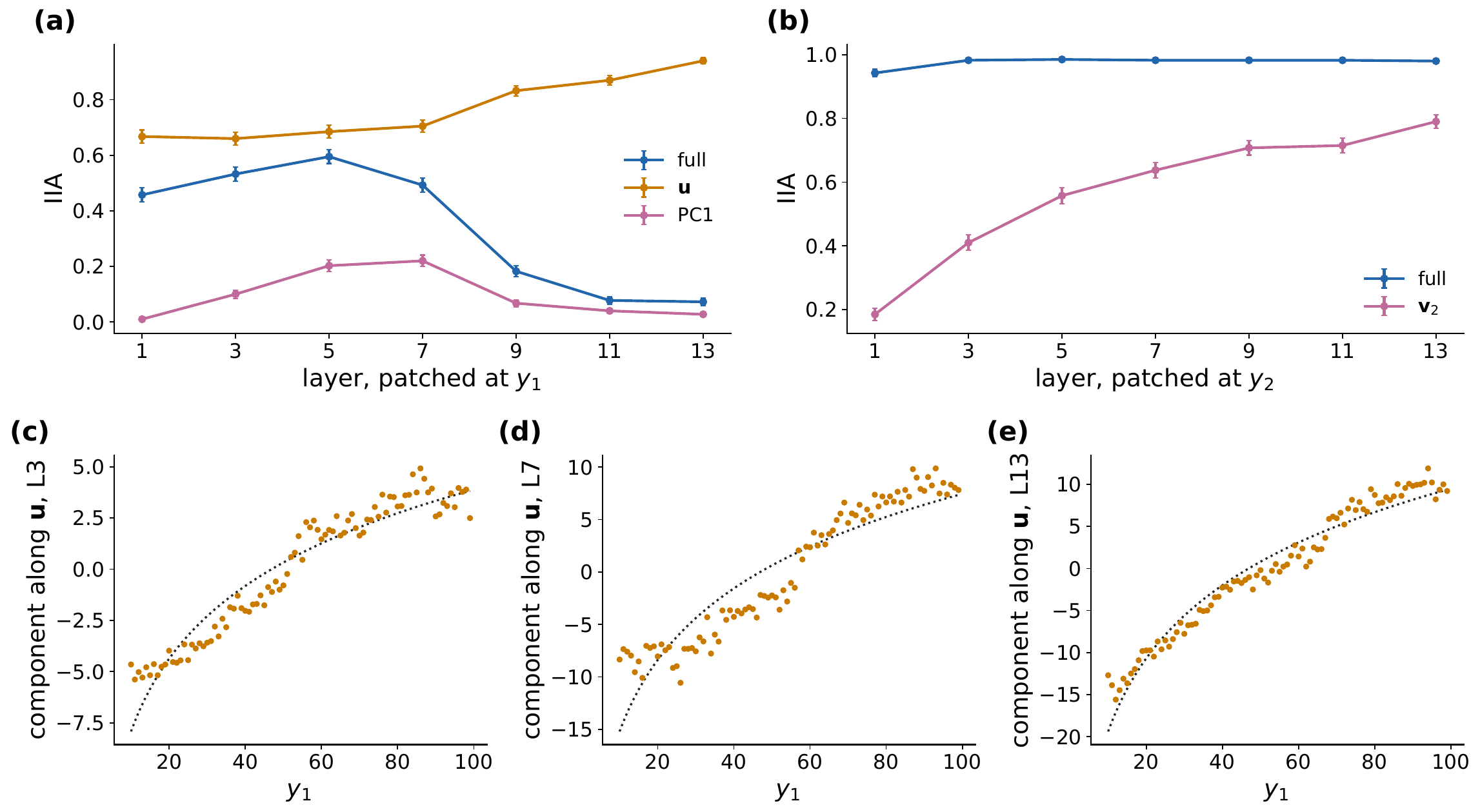}
  \caption{The number-representation analysis at layers $1$ to $13$, with $\mathbf{u}$ refitted at each layer. (a)~IIA at the position of $y_1$, with $y_1$ perturbed, for a full-rank patch, $\mathbf{u}$ and that layer's top principal component. (b)~IIA at the position of $y_2$, with $y_2$ perturbed, for a full-rank patch and $\mathbf{v}_2$, taken as that layer's top principal component. (c--e)~Component along $\mathbf{u}$ against $y_1$ at layers 3, 7 and 13, as per-value means, with the fit $p\log y_1+q$ (dotted).}
  \label{fig:app-u-layers}
\end{figure}

\begin{figure}[!ht]
  \centering
  \includegraphics[width=\linewidth]{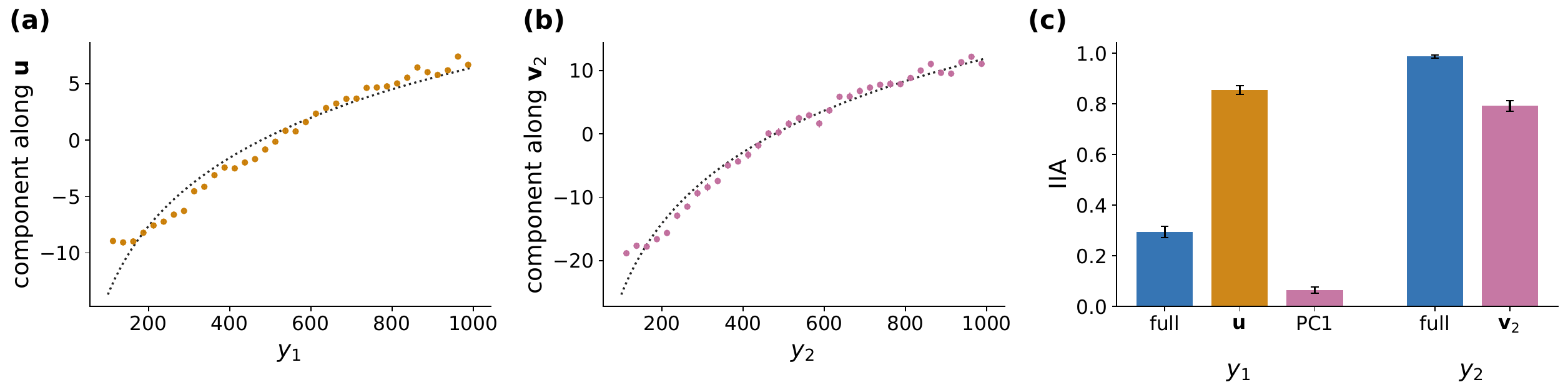}
  \caption{The number representations with three-digit operands. (a)~Component along $\mathbf{u}$ at the position of $y_1$ against $y_1$, and (b)~component along $\mathbf{v}_2$ at the position of $y_2$ against $y_2$, both averaged in bins of $25$, with the fit $p\log y+q$ (dotted). (c)~IIA of a full-rank patch, $\mathbf{u}$ and PC1 at the position of $y_1$ with $y_1$ perturbed, and of a full-rank patch and $\mathbf{v}_2$ at the position of $y_2$ with $y_2$ perturbed.}
  \label{fig:app-three-digit}
\end{figure}

\begin{figure}[!ht]
  \centering
  \includegraphics[width=\linewidth]{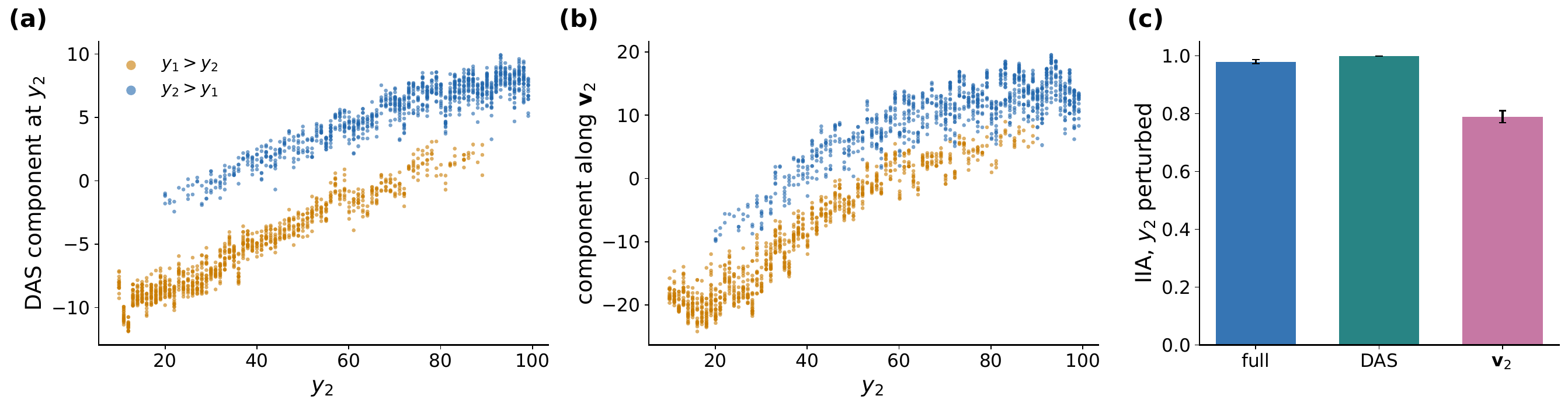}
  \caption{The layer-13 residual at the position of $y_2$ on the $2{,}000$-prompt cloud, coloured by which number is larger. (a)~Component along a rank-one direction fitted with $y_2$ perturbed, against $y_2$. (b)~Component along $\mathbf{v}_2$, the top principal component, against $y_2$. (c)~IIA with $y_2$ perturbed for a full-rank patch, the fitted direction (DAS) and $\mathbf{v}_2$.}
  \label{fig:app-v2-shortcut}
\end{figure}

\begin{figure}[!ht]
  \centering
  \includegraphics[width=\linewidth]{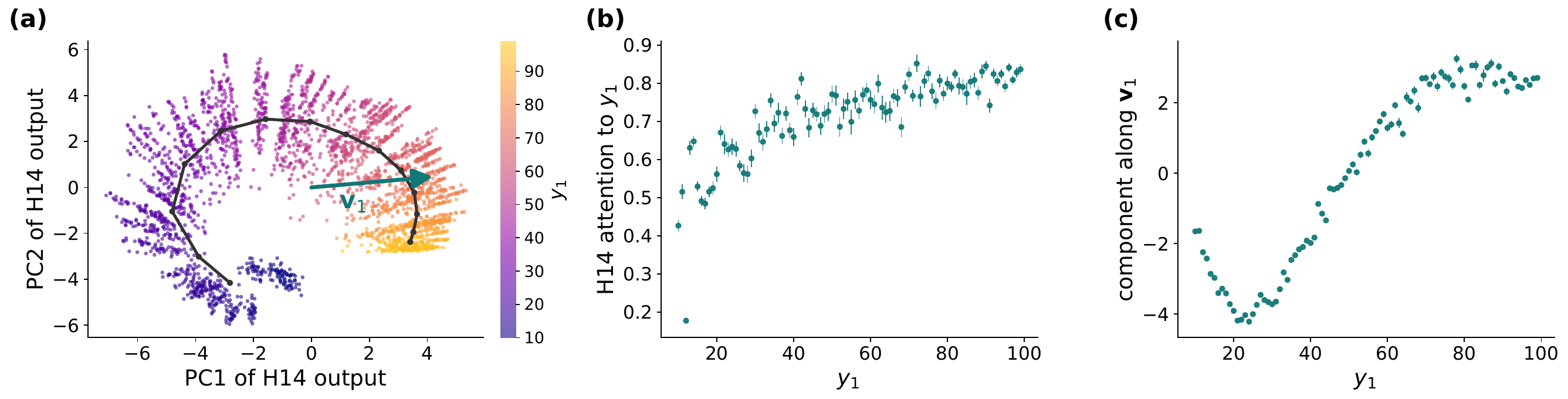}
  \caption{The output of head L14.H14 at the position of $y_2$. (a)~Its top two principal components, coloured by $y_1$, with the smoothed mean position along $y_1$ (line) and the direction of $\mathbf{v}_1$'s projection onto the plane (arrow). (b)~The head's attention from the position of $y_2$ to $y_1$, against $y_1$, as per-value means. (c)~Component along $\mathbf{v}_1$ against $y_1$, as per-value means.}
  \label{fig:app-transport}
\end{figure}

\FloatBarrier
\subsection{The shared representation}
\label{app:res-shared}

Probes and position recovery complement the main-text IIA results for the shared representation
at the position of $y_2$.

\begin{figure}[!ht]
  \centering
  \includegraphics[width=0.8\linewidth]{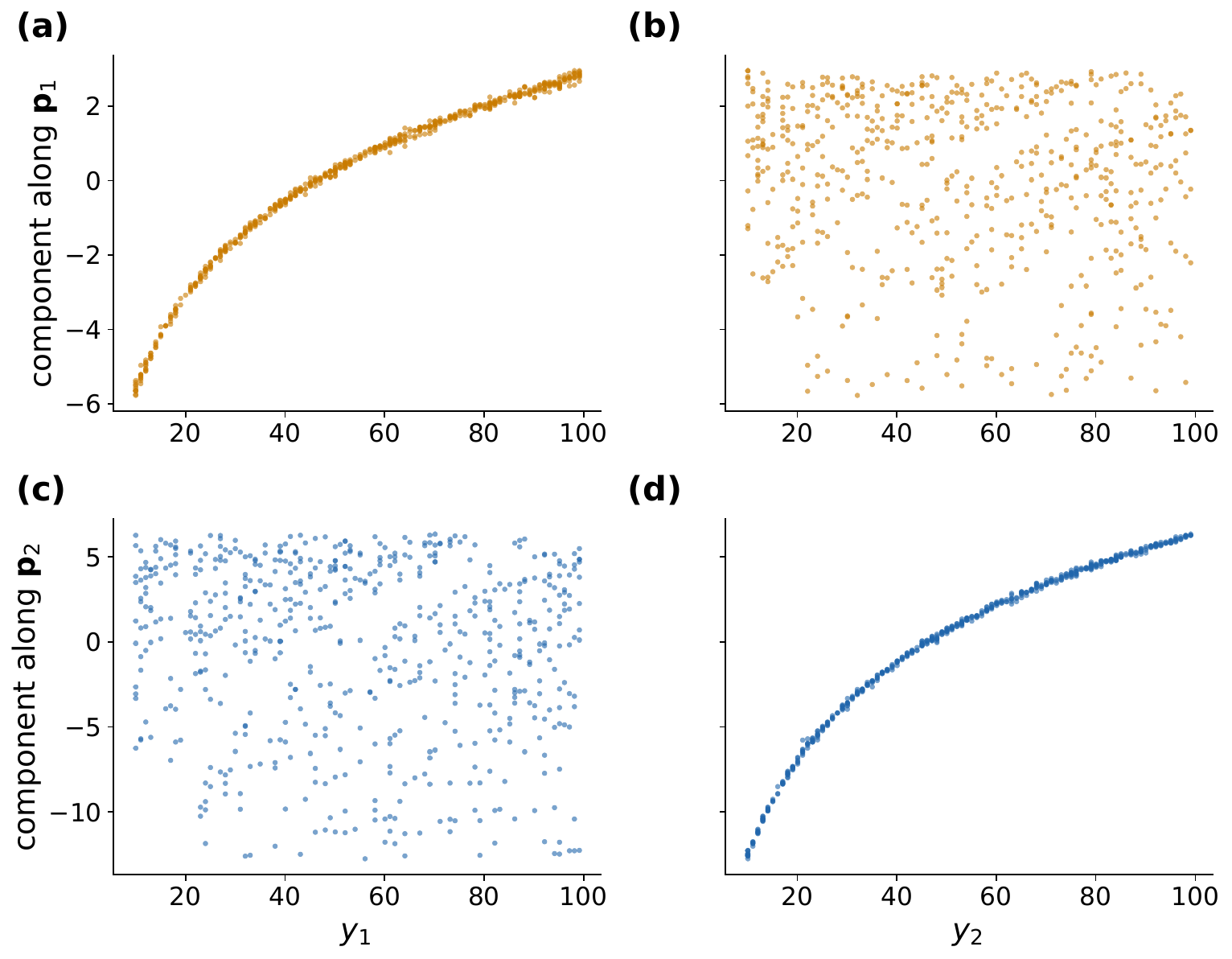}
  \caption{Components of the pre-MLP layer-14 residual at the position of $y_2$ along ridge probes for $\log y_1$ ($\mathbf{p}_1$, top row) and $\log y_2$ ($\mathbf{p}_2$, bottom row), against $y_1$ (left) and $y_2$ (right), on $600$ held-out prompts.}
  \label{fig:app-shared-probes}
\end{figure}

\begin{figure}[!ht]
  \centering
  \includegraphics[width=0.75\linewidth]{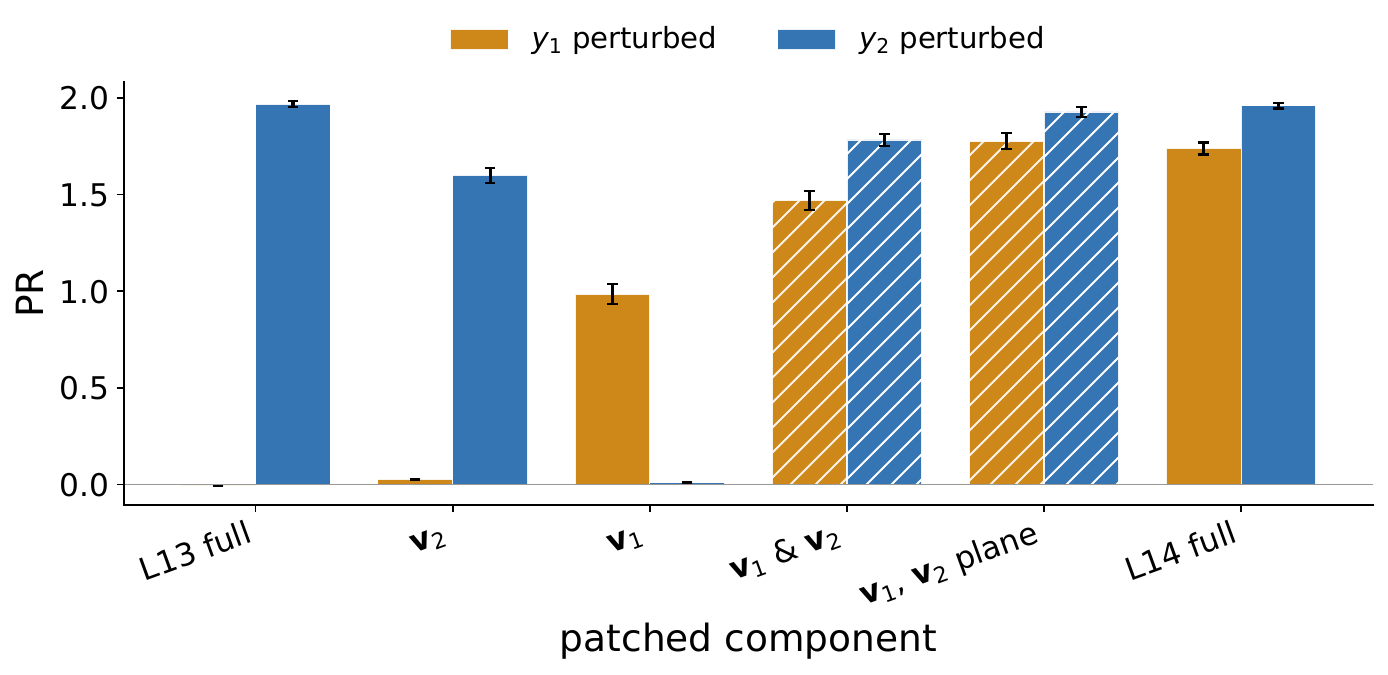}
  \caption{Position recovery at the position of $y_2$ for the patches of the main-text shared-representation figure, with $y_1$ or $y_2$ perturbed: the full layer-13 residual, $\mathbf{v}_2$, $\mathbf{v}_1$, $\mathbf{v}_1$ and $\mathbf{v}_2$ as two separate rank-one patches, the rank-two $(\mathbf{v}_1,\mathbf{v}_2)$ plane in the pre-MLP layer-14 residual, and the full pre-MLP layer-14 residual. Hatched bars are the two conditions that patch both directions.}
  \label{fig:app-shared-posrec}
\end{figure}

\FloatBarrier
\subsection{The comparator neurons}
\label{app:res-neurons}

Freezing, attribution, receptive fields and connectivity for MLP14 and MLP15 appear in the order of
the selection procedure of Sections~\ref{app:methods-attribution}--\ref{app:methods-connect}.

\begin{figure}[!ht]
  \centering
  \includegraphics[width=\linewidth]{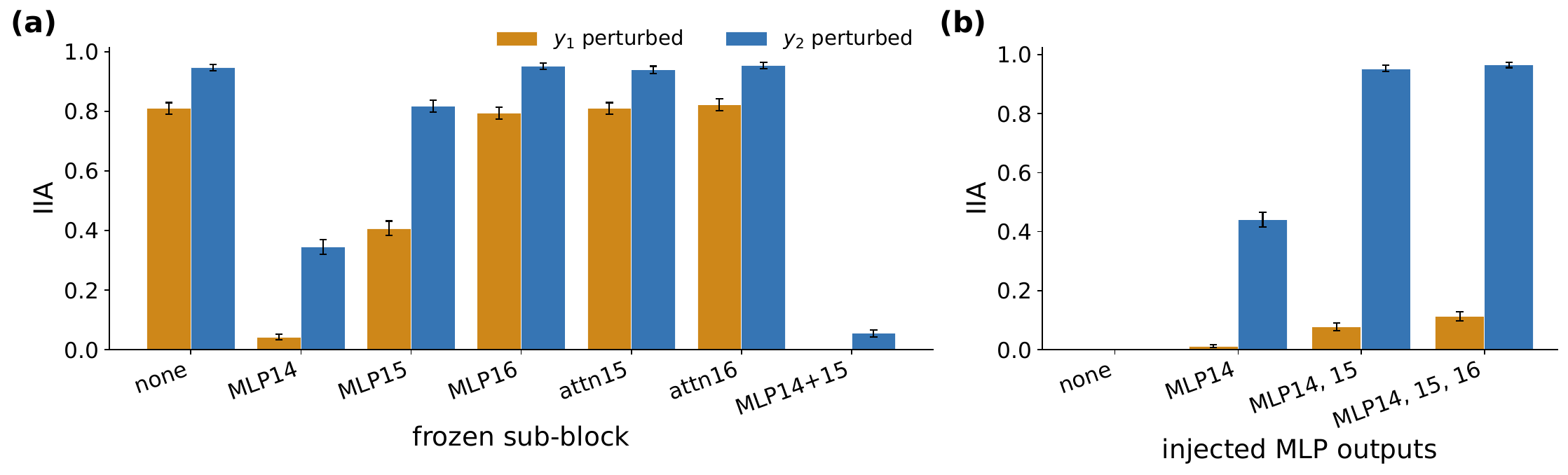}
  \caption{(a)~IIA of the rank-two $(\mathbf{v}_1,\mathbf{v}_2)$ plane patch when one sub-block's output at the position of $y_2$ is frozen to its corrupted value, with $y_1$ or $y_2$ perturbed. The layer-14 attention freeze is omitted, because the pre-MLP patch is written through that output (Section~\ref{app:methods-sites}). (b)~IIA when only the clean outputs of the listed MLPs are injected at the position of $y_2$.}
  \label{fig:app-mlp-freeze}
\end{figure}

\begin{figure}[!ht]
  \centering
  \includegraphics[width=\linewidth]{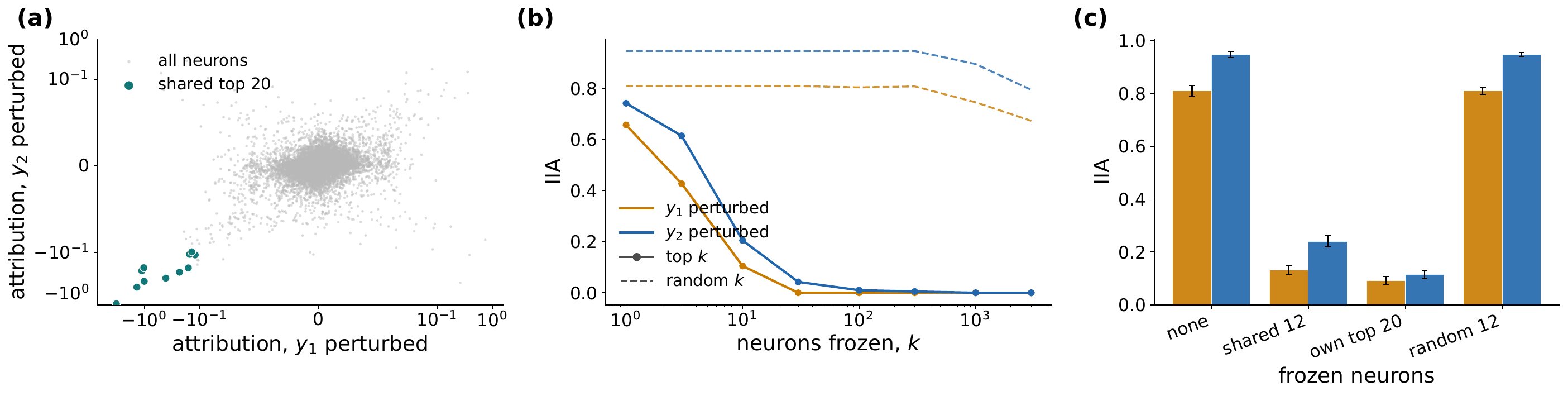}
  \caption{(a)~Attribution scores of the $37{,}888$ neurons of MLP14 and MLP15, with $y_1$ perturbed ($x$-axis) and $y_2$ perturbed ($y$-axis), on symmetric-log axes. The $12$ neurons in the top $20$ of both cases are highlighted. (b)~IIA of the plane patch when the top $k$ neurons by attribution (solid) or $k$ random neurons (dashed) are frozen. (c)~IIA with no neurons frozen, with the $12$ shared neurons frozen, with each case's own $20$ top-ranked neurons frozen, and with $12$ random neurons frozen.}
  \label{fig:app-attribution}
\end{figure}

\begin{figure}[!ht]
  \centering
  \includegraphics[width=\linewidth]{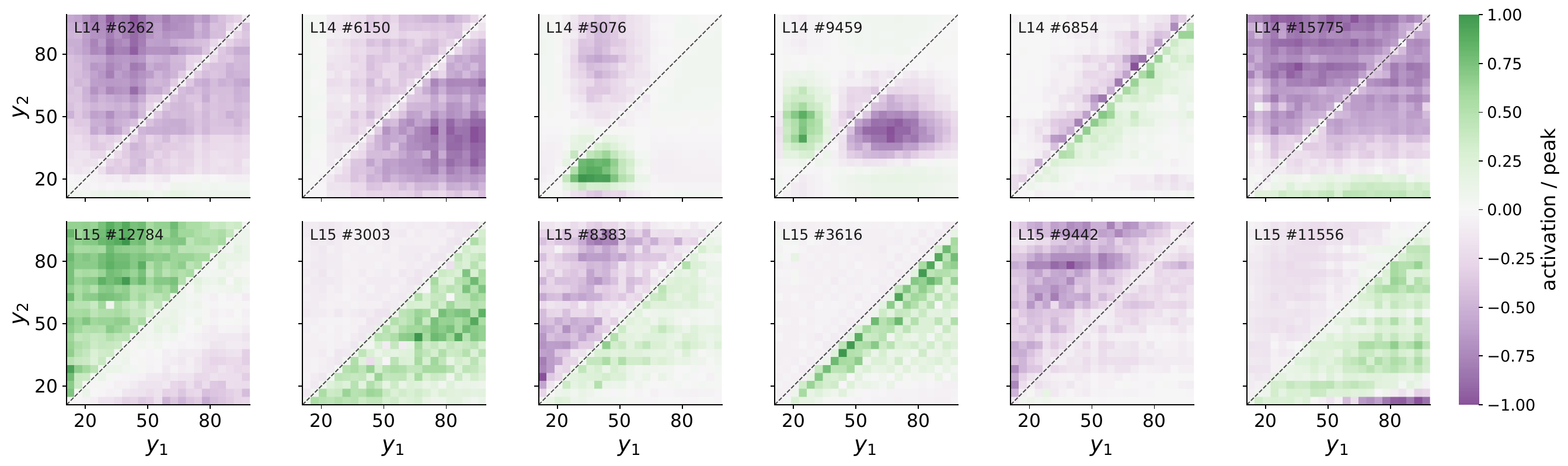}
  \caption{Receptive fields of the $12$ shared neurons, six in MLP14 (top) and six in MLP15 (bottom), ordered by worst-case attribution rank. Each field is the post-SwiGLU activation at the position of $y_2$ over a grid of $(y_1,y_2)$ prompts with stride $4$, divided by its peak. The dashed line is $y_1=y_2$.}
  \label{fig:app-rf-shared}
\end{figure}

\begin{figure}[!ht]
  \centering
  \includegraphics[width=\linewidth]{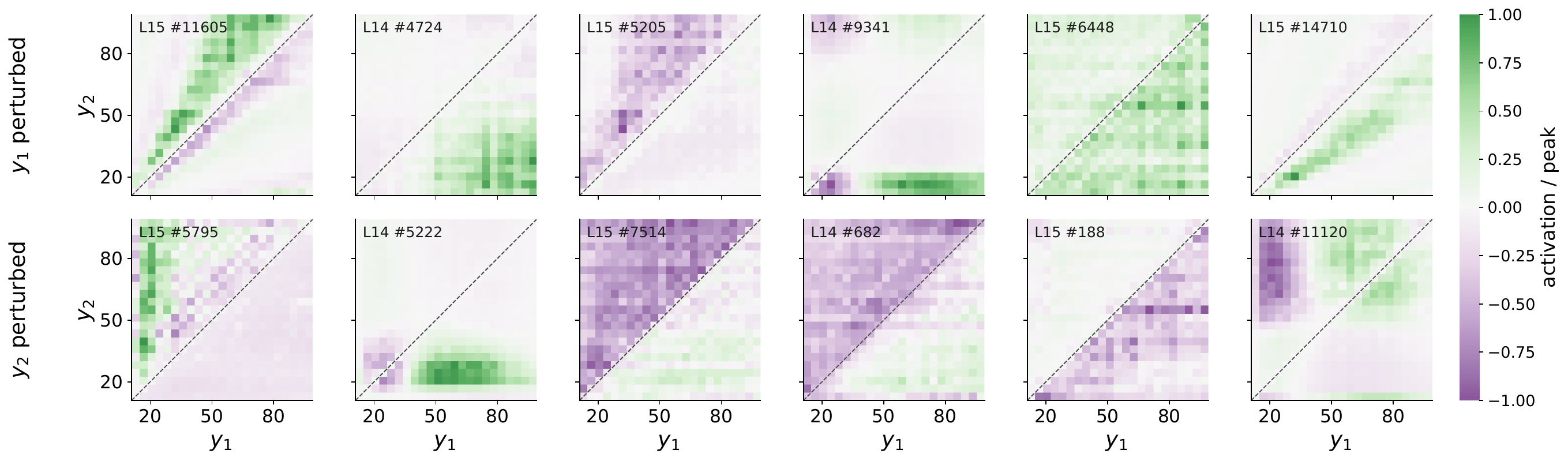}
  \caption{Receptive fields, drawn as in Figure~\ref{fig:app-rf-shared}, of the six highest-ranked neurons that are in the top $20$ of one perturbation case only: $y_1$ perturbed (top) and $y_2$ perturbed (bottom).}
  \label{fig:app-rf-exclusive}
\end{figure}

\begin{figure}[!ht]
  \centering
  \includegraphics[width=\linewidth]{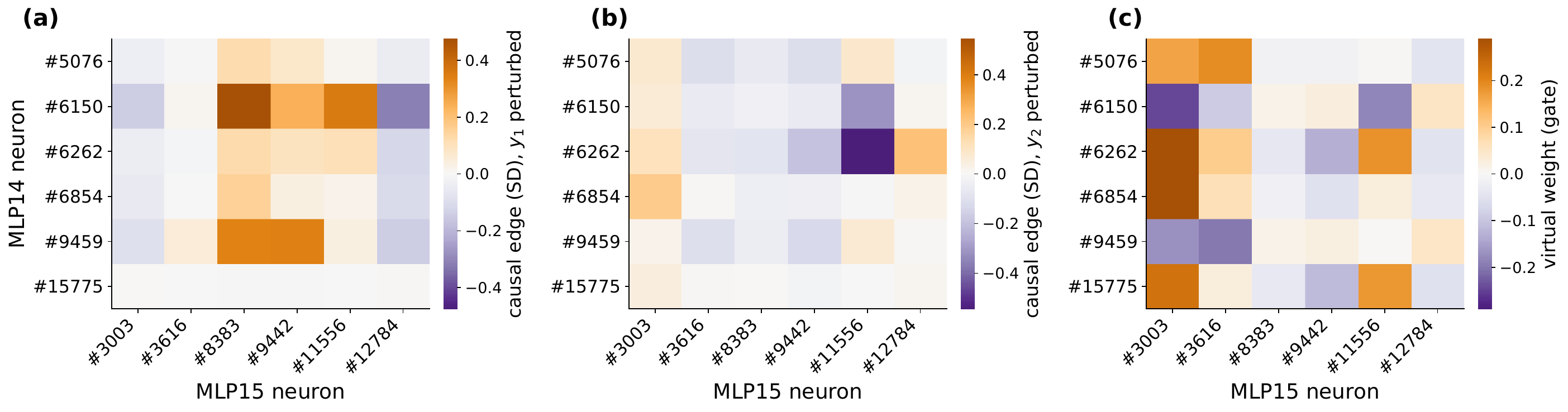}
  \caption{(a,b)~Causal edges from each shared MLP14 neuron (rows) to each shared MLP15 neuron (columns): the shift in the MLP15 neuron's activation, in units of its standard deviation, when only the MLP14 neuron is set to its clean value, with (a)~$y_1$ or (b)~$y_2$ perturbed. (c)~Virtual weights through the gate projection, computed as the cosine between each MLP14 neuron's write direction and each MLP15 neuron's normalization-folded gate read direction (Section~\ref{app:methods-connect}).}
  \label{fig:app-connectivity}
\end{figure}

\begin{figure}[!ht]
  \centering
  \includegraphics[width=\linewidth]{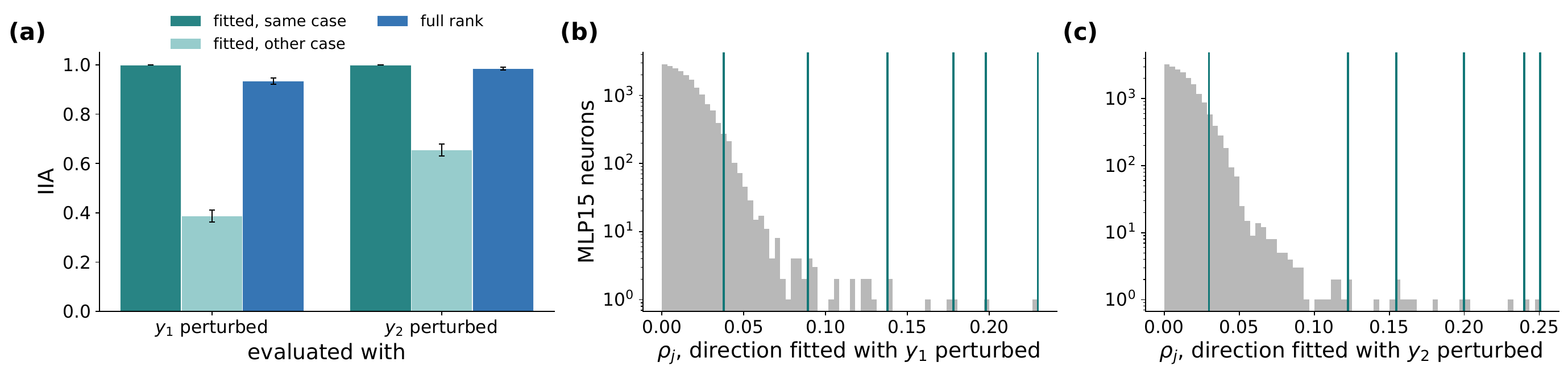}
  \caption{(a)~IIA of a rank-one direction in the layer-15 residual at the position of $y_2$, fitted in the case it is evaluated on (fitted, same case) or in the other case (fitted, other case), and of the full-rank layer-15 patch, for each perturbation case. (b,c)~Histograms of the alignment $\rho_j$ of the write direction of every MLP15 neuron with the direction fitted with (b)~$y_1$ or (c)~$y_2$ perturbed, on a log count axis. Vertical lines mark the six shared MLP15 neurons.}
  \label{fig:app-l15-direction}
\end{figure}

\FloatBarrier
\subsection{Three numbers}
\label{app:res-three}

A summary figure covers the positions of $y_3$ and $y_2$ and the last token, and the figures after
it expand each part.

\begin{figure}[!ht]
  \centering
  \includegraphics[width=\linewidth]{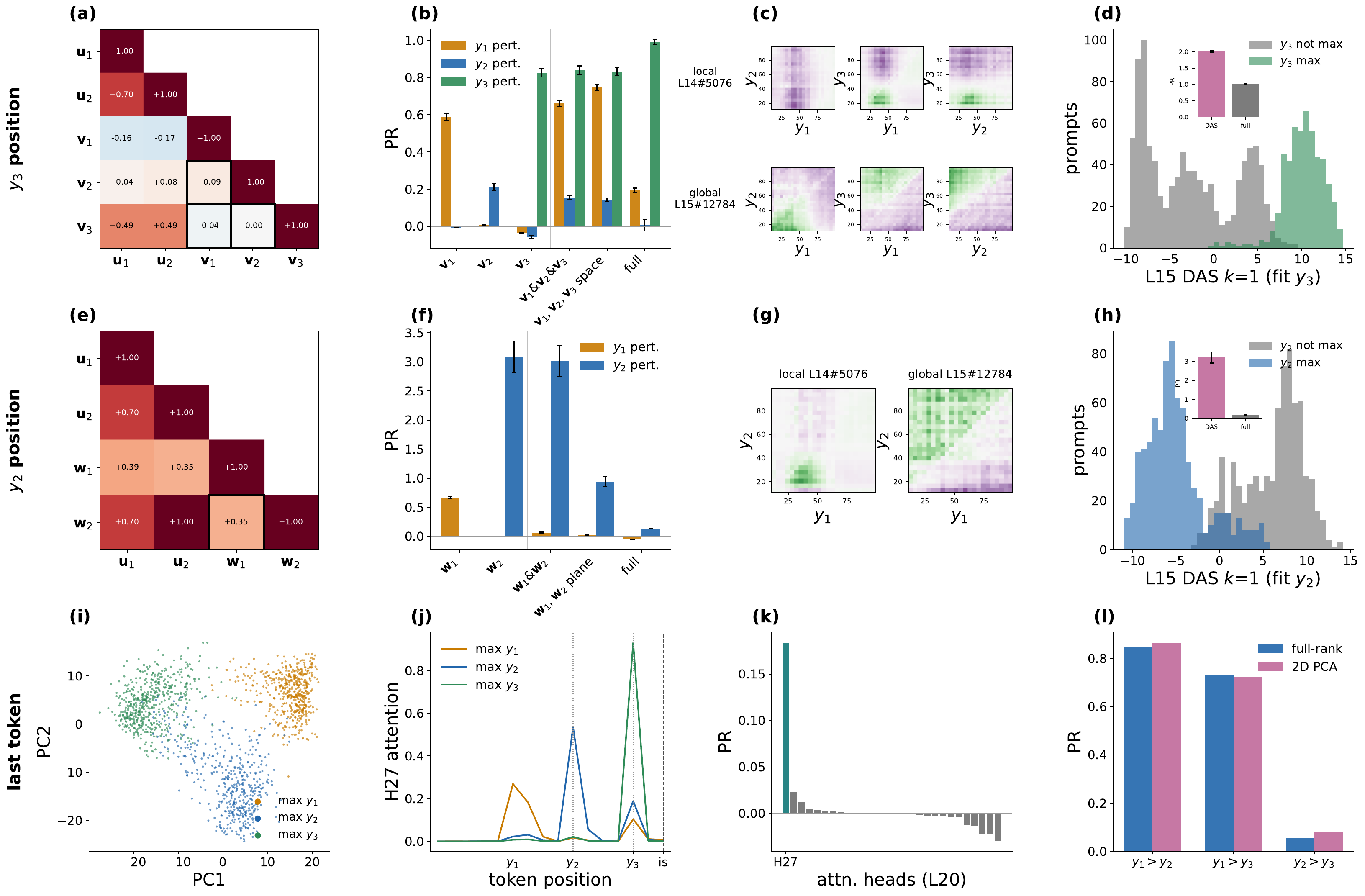}
  \caption{The three-number analysis at the position of $y_3$ (top row), the position of $y_2$ (middle row) and the last token (bottom row). (a)~Cosine similarities between $\mathbf{u}_1$, $\mathbf{u}_2$, $\mathbf{v}_1$, $\mathbf{v}_2$ and $\mathbf{v}_3$; the pairs among $\mathbf{v}_1,\mathbf{v}_2,\mathbf{v}_3$ are boxed. (b)~Position recovery at the position of $y_3$, by perturbed number, for each direction alone, the three as separate rank-one patches, their span in the pre-MLP layer-14 residual, and the full layer-14 residual. (c)~Receptive fields of L14\#5076 (top) and L15\#12784 (bottom) at the position of $y_3$ over a $(y_1,y_2,y_3)$ grid with stride $4$, shown as the three pairwise marginals, each averaged over the third number and divided by its own peak. (d)~The $1{,}500$-prompt cloud projected on a rank-one direction in the layer-15 residual fitted with $y_3$ perturbed, split by whether $y_3$ is the maximum; the inset shows the position recovery of this direction and of the full layer-15 residual. (e--h)~The same at the position of $y_2$, where $\mathbf{w}_1$ and $\mathbf{w}_2$ are rank-one directions in the layer-13 residual fitted with $y_1$ and $y_2$ perturbed ($\mathbf{w}_2$ and $\mathbf{u}_2$ are the same fit). The fields in (g) are over $(y_1,y_2)$ with $y_3=55$. (i)~Top two principal components of the residual leaving layer 20 at the last token, coloured by which operand holds the maximum. (j)~Attention of head L20.H27 from the last token to the question tokens, averaged by the position of the maximum. (k)~Position recovery of each layer-20 head's own output at the last token, for the order $y_1>y_2>y_3$. (l)~Position recovery of a full-rank patch and of a patch of the top two principal components of the layer-20 residual at the last token, for three value orders; the principal components are fitted on a separate $1{,}500$-prompt cloud.}
  \label{fig:app-k3-summary}
\end{figure}

\begin{figure}[!ht]
  \centering
  \includegraphics[width=\linewidth]{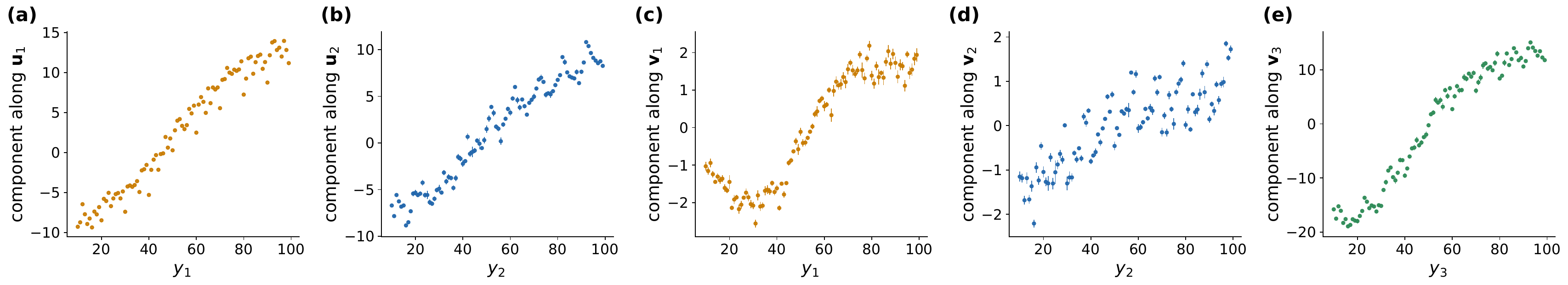}
  \caption{Component along each direction of the three-number analysis against the number it carries, as per-value means on the $1{,}500$-prompt cloud. (a,b)~$\mathbf{u}_1$ and $\mathbf{u}_2$ in the layer-13 residual at the positions of $y_1$ and $y_2$. (c,d)~$\mathbf{v}_1$ and $\mathbf{v}_2$ in the outputs of heads L14.H14 and L14.H18 at the position of $y_3$. (e)~$\mathbf{v}_3$ in the layer-13 residual at the position of $y_3$.}
  \label{fig:app-k3-directions}
\end{figure}

\begin{figure}[!ht]
  \centering
  \includegraphics[width=\linewidth]{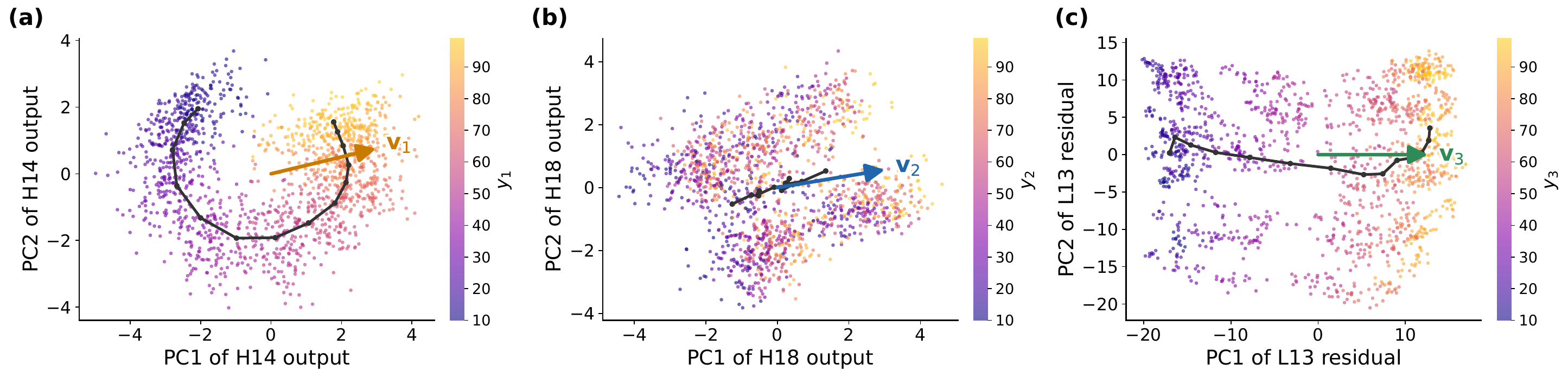}
  \caption{The spaces at the position of $y_3$ that contain the three-number directions, each in its top two principal components, coloured by the number its direction carries, with the smoothed mean position along that number (line) and the direction of the corresponding projection onto the plane (arrow). (a)~The output of head L14.H14 with $\mathbf{v}_1$. (b)~The output of head L14.H18 with $\mathbf{v}_2$. (c)~The layer-13 residual with $\mathbf{v}_3$.}
  \label{fig:app-k3-manifolds}
\end{figure}

\begin{figure}[!ht]
  \centering
  \includegraphics[width=0.55\linewidth]{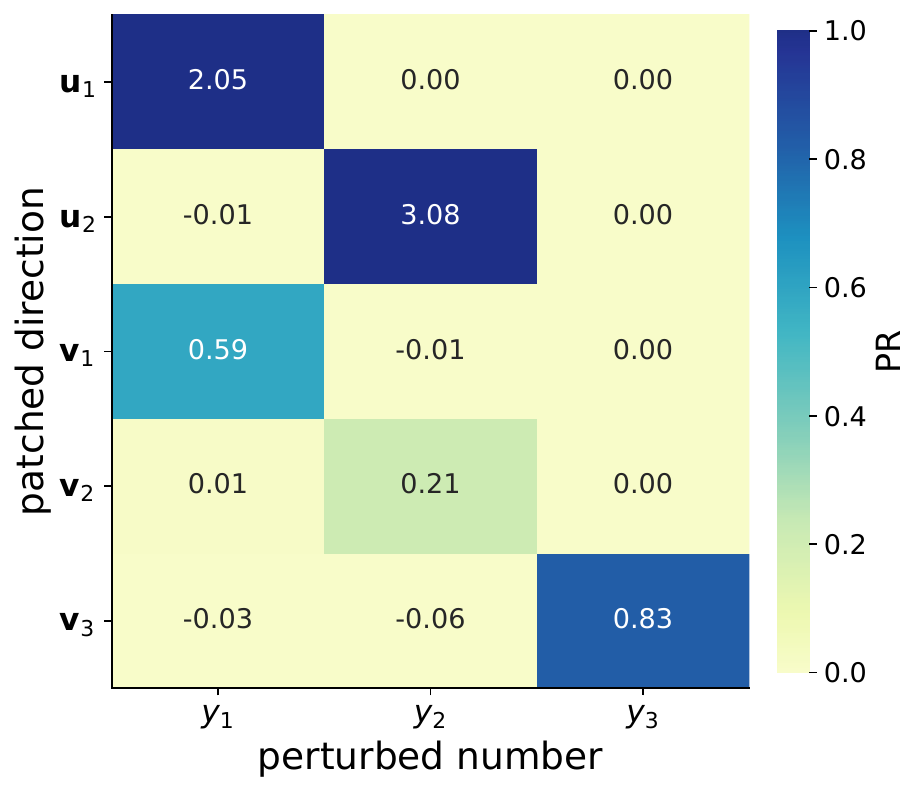}
  \caption{Position recovery of a rank-one patch of each direction (rows) in each perturbation case (columns). Colours are clipped at $1$; printed values are not.}
  \label{fig:app-k3-dissociation}
\end{figure}

\begin{figure}[!ht]
  \centering
  \includegraphics[width=0.8\linewidth]{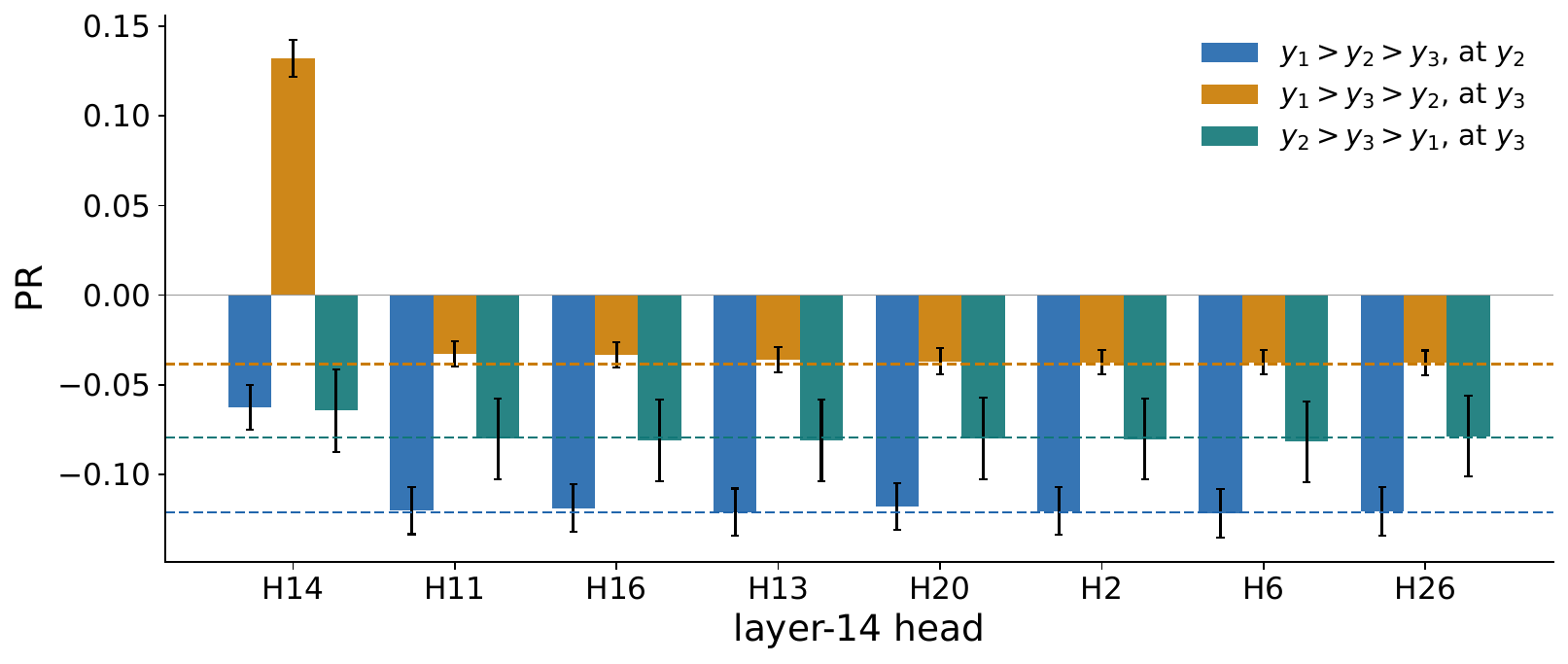}
  \caption{Position recovery of each layer-14 head's own output, interchanged at the runner-up's position together with the full layer-13 residual there, for three value orders and the eight heads with the largest mean absolute effect. Dashed lines show the layer-13 patch alone for each order.}
  \label{fig:app-k3-heads}
\end{figure}

\begin{figure}[!ht]
  \centering
  \includegraphics[width=\linewidth]{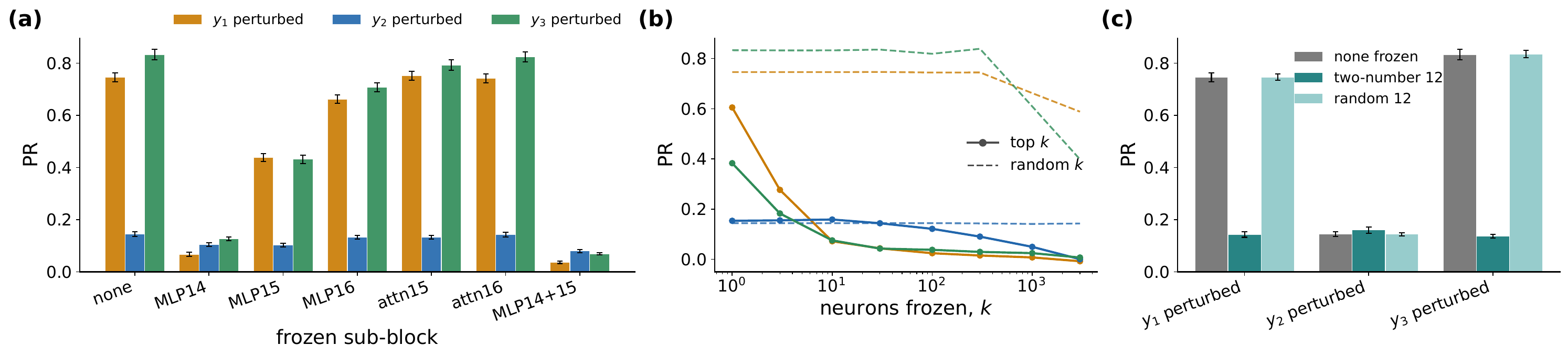}
  \caption{Position recovery of the patch of the span of $\mathbf{v}_1,\mathbf{v}_2,\mathbf{v}_3$ in the pre-MLP layer-14 residual at the position of $y_3$, by perturbed number, while parts of the network are frozen. (a)~One sub-block's output at the position of $y_3$ frozen to its corrupted value. (b)~The top $k$ neurons by this task's attribution ranking (solid) or $k$ random neurons (dashed) frozen. (c)~No neurons frozen, the $12$ shared neurons of the two-number task frozen, and $12$ random neurons frozen.}
  \label{fig:app-k3-neurons}
\end{figure}

\begin{figure}[!ht]
  \centering
  \includegraphics[width=0.85\linewidth]{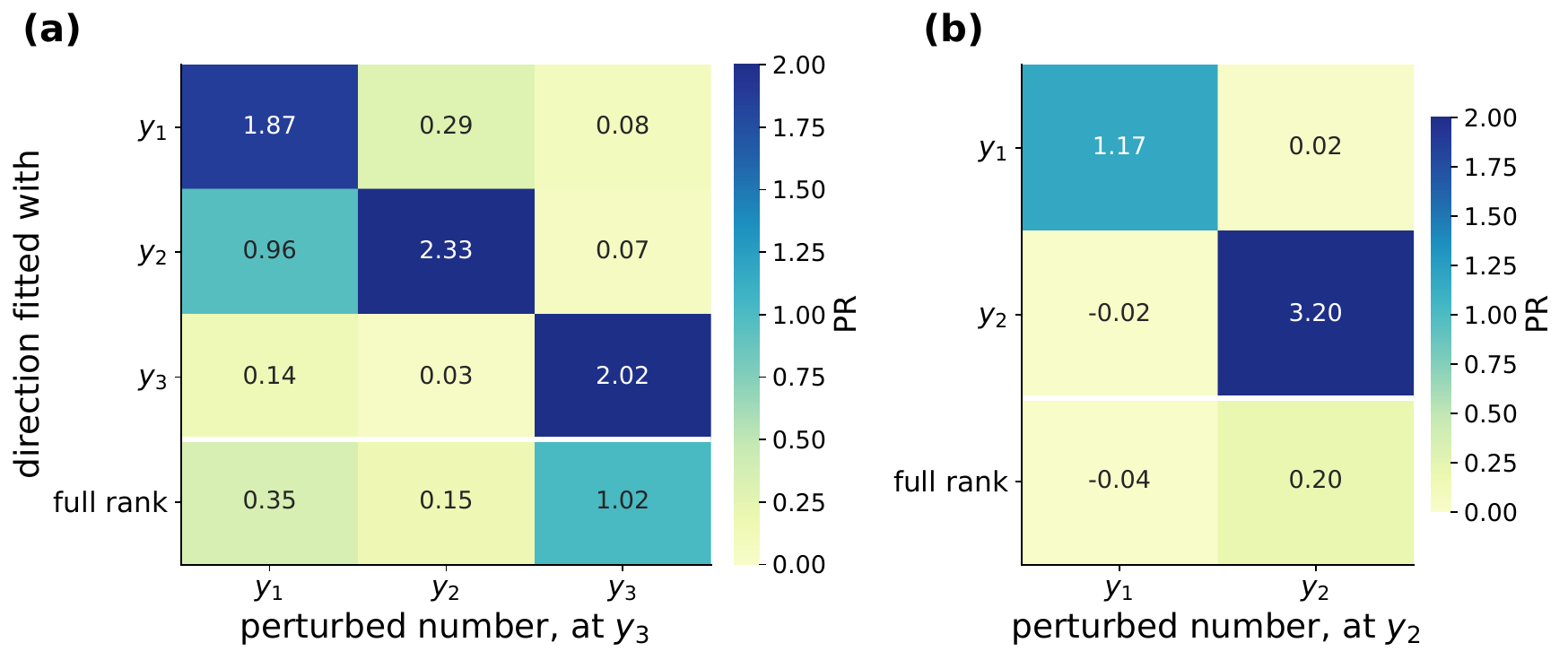}
  \caption{Position recovery of a rank-one direction in the layer-15 residual fitted in one case (rows) and evaluated in each case (columns), with the full-rank layer-15 patch in the last row, (a)~at the position of $y_3$ and (b)~at the position of $y_2$. Colours are clipped at $2$; printed values are not.}
  \label{fig:app-k3-l15}
\end{figure}

\begin{figure}[!ht]
  \centering
  \includegraphics[width=\linewidth]{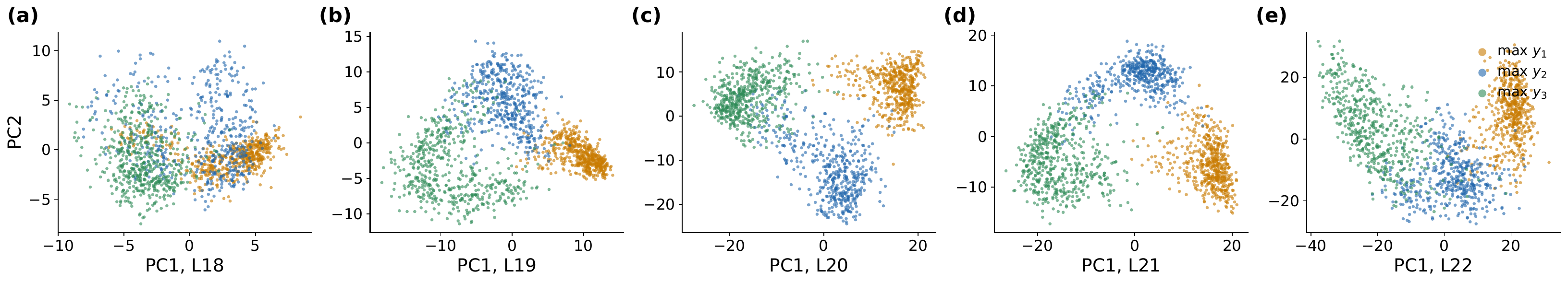}
  \caption{Top two principal components of the residual stream leaving layers 18 to 22 at the last token, on $1{,}500$ clean three-number prompts, coloured by which operand holds the maximum.}
  \label{fig:app-k3-readout-pca}
\end{figure}

\begin{figure}[!ht]
  \centering
  \includegraphics[width=\linewidth]{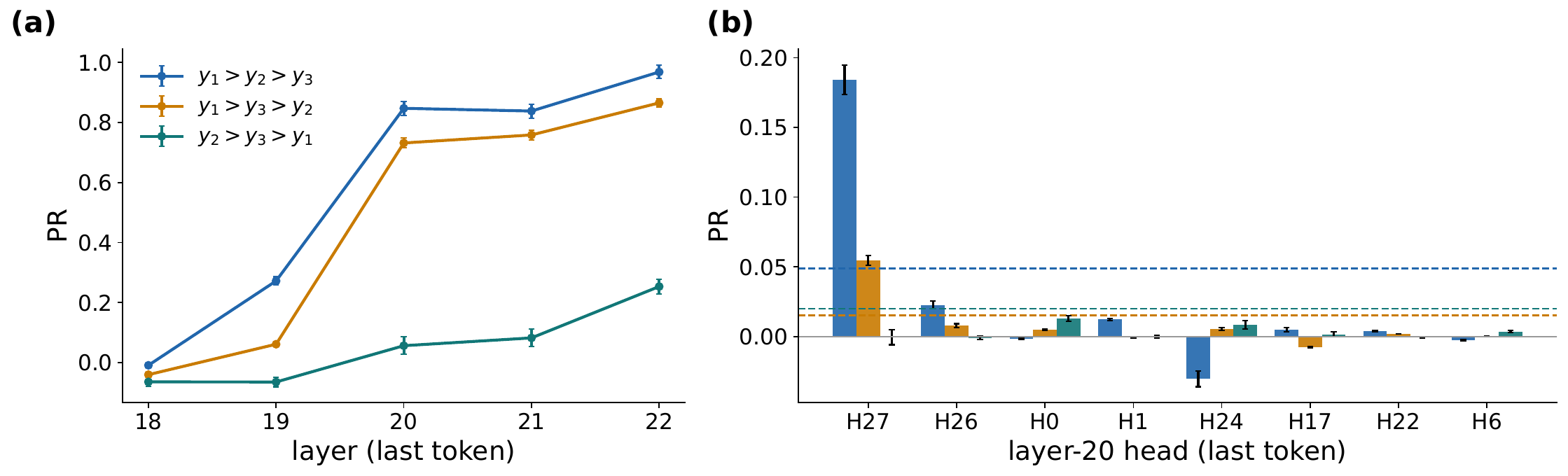}
  \caption{(a)~Position recovery of a full-rank patch of the residual leaving each of layers 18 to 22 at the last token, for three value orders. (b)~Position recovery of each layer-20 head's own output at the last token, for the eight heads with the largest effect and the same three orders. Dashed lines show a patch of the whole layer-20 attention sublayer for each order (Section~\ref{app:methods-readout}).}
  \label{fig:app-k3-readout-causal}
\end{figure}

\begin{figure}[!ht]
  \centering
  \includegraphics[width=0.55\linewidth]{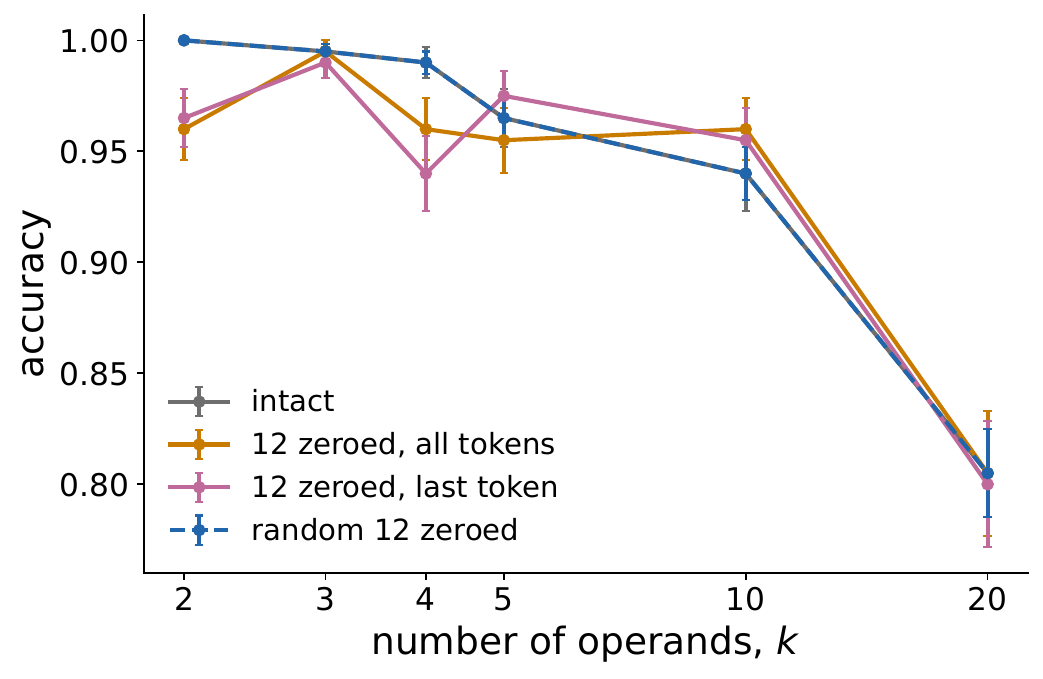}
  \caption{Accuracy on $\max(y_1,\dots,y_k)$ for $200$ two-digit prompts per length, scored on the full decoded number (Section~\ref{app:methods-metrics}): the intact model, the $12$ shared neurons zeroed at every token, the same neurons zeroed only at the last token of the final operand (``last token'' in the legend), and $12$ random neurons zeroed at every token (two random sets, pooled).}
  \label{fig:app-ablation}
\end{figure}

\FloatBarrier

\subsection{IIA on the whole generated number}
\label{app:res-whole}

We rescore the main-text interventions on the whole number the model generates.

\begin{figure}[!ht]
  \centering
  \includegraphics[width=\linewidth]{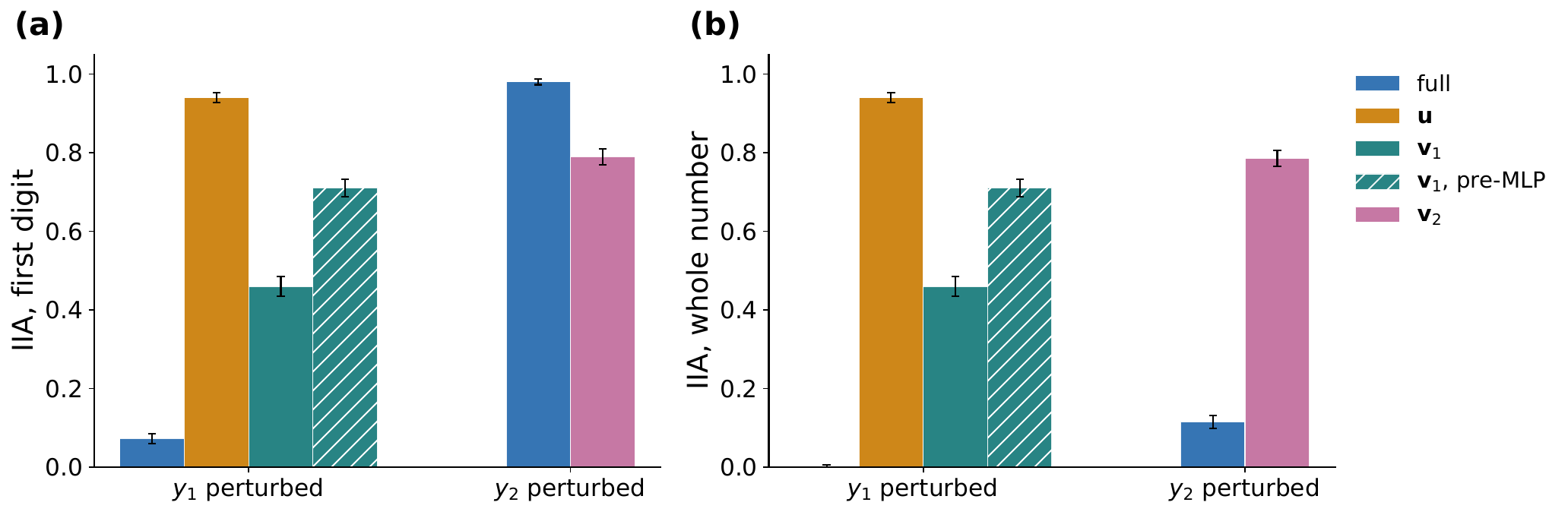}
  \caption{IIA of the number-representation patches, scored on (a)~the first generated token, the leading digit of $r$ (Section~\ref{app:methods-metrics}), and (b)~the whole generated number: the answer decoded greedily for one token more than the operands' digit count, counted correct when its first integer equals $r$. With $y_1$ perturbed: the full layer-13 residual and $\mathbf{u}$ at the position of $y_1$, and $\mathbf{v}_1$ at the position of $y_2$, patched inside the output of L14.H14 as in the main text or in the pre-MLP layer-14 residual (hatched). With $y_2$ perturbed: the full layer-13 residual and $\mathbf{v}_2$ at the position of $y_2$. The rank-one patches score the same on both, while the full-rank patches mostly generate a number other than $r$ that starts with its leading digit.}
  \label{fig:app-whole-directions}
\end{figure}

\begin{figure}[!ht]
  \centering
  \includegraphics[width=\linewidth]{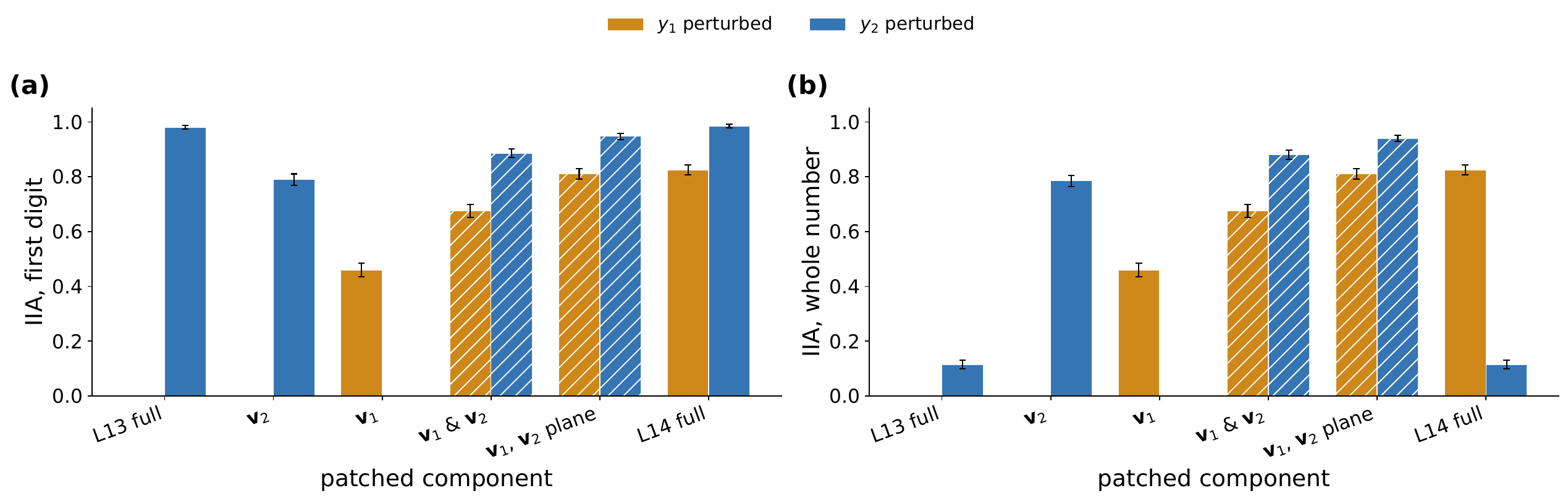}
  \caption{IIA of the patches of the main-text shared-representation figure at the position of $y_2$, with $y_1$ or $y_2$ perturbed, scored on (a)~the first generated token and (b)~the whole generated number, as in Figure~\ref{fig:app-whole-directions}: the full layer-13 residual, $\mathbf{v}_2$, $\mathbf{v}_1$, $\mathbf{v}_1$ and $\mathbf{v}_2$ as two separate rank-one patches, the rank-two $(\mathbf{v}_1,\mathbf{v}_2)$ plane in the pre-MLP layer-14 residual, and the full residual leaving layer 14. Hatched bars are the two conditions that patch both directions. Only the full-rank patches with $y_2$ perturbed lose IIA on the whole number.}
  \label{fig:app-whole-shared}
\end{figure}

\begin{figure}[!ht]
  \centering
  \includegraphics[width=0.45\linewidth]{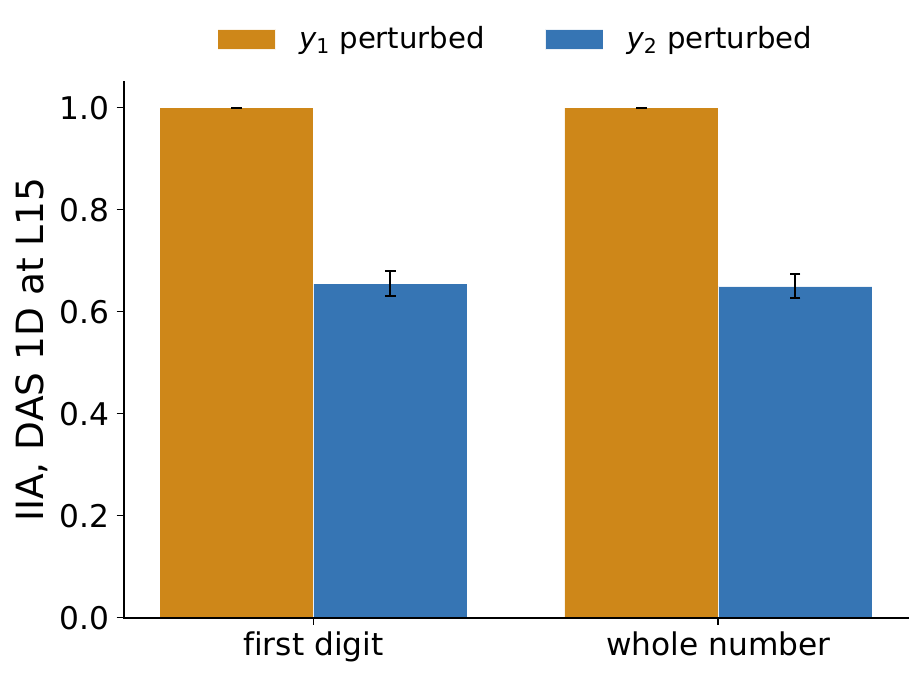}
  \caption{IIA of the rank-one direction in the layer-15 residual at the position of $y_2$, fitted with $y_1$ perturbed and evaluated in both cases, scored on the first generated token and on the whole generated number, as in Figure~\ref{fig:app-whole-directions}.}
  \label{fig:app-whole-l15}
\end{figure}

\begin{table}[!ht]
  \centering
  \small
  \caption{IIA on the first generated token and on the whole generated number for the patches of Figures~\ref{fig:app-whole-directions}--\ref{fig:app-whole-l15}. Dashes mark cases in which a patch is not evaluated.}
  \label{tab:app-whole-iia}
  \begin{tabular}{lcccc}
\toprule
 & \multicolumn{2}{c}{$y_1$ perturbed} & \multicolumn{2}{c}{$y_2$ perturbed} \\
\cmidrule(lr){2-3}\cmidrule(lr){4-5}
patch & first digit & whole number & first digit & whole number \\
\midrule
L13 full at $y_1$ & 0.072 & 0.003 & -- & -- \\
$\mathbf{u}$ & 0.940 & 0.940 & -- & -- \\
$\mathbf{v}_1$ & 0.460 & 0.460 & 0.000 & 0.000 \\
$\mathbf{v}_1$, pre-MLP & 0.710 & 0.710 & -- & -- \\
L13 full at $y_2$ & 0.000 & 0.000 & 0.980 & 0.115 \\
$\mathbf{v}_2$ & 0.000 & 0.000 & 0.790 & 0.785 \\
$\mathbf{v}_1$ \& $\mathbf{v}_2$ & 0.675 & 0.675 & 0.885 & 0.880 \\
$(\mathbf{v}_1, \mathbf{v}_2)$ plane & 0.810 & 0.810 & 0.948 & 0.940 \\
L14 full & 0.825 & 0.825 & 0.985 & 0.115 \\
L15 DAS & 1.000 & 1.000 & 0.655 & 0.650 \\
\bottomrule
\end{tabular}

\end{table}

\begin{table}[!ht]
  \centering
  \small
  \caption{Generated answers under the patches of Figure~\ref{fig:app-whole-directions}, five per patch, chosen to include correct answers and mistakes: the perturbed number, the clean and corrupted operands, $r$, and the generated number. A patch succeeds when the model generates $r$.}
  \label{tab:app-whole-examples-a}
  \begin{tabular}{llcccc}
\toprule
patch & perturbed & clean $(y_1, y_2)$ & corrupted $(y_1, y_2)$ & $r$ & generated \\
\midrule
L13 full at $y_1$ & $y_1$ & (93, 41) & (13, 41) & 13 & 13 \\
 & $y_1$ & (96, 46) & (20, 46) & 20 & 26 \\
 & $y_1$ & (69, 42) & (13, 42) & 13 & 12 \\
 & $y_1$ & (60, 38) & (15, 38) & 15 & 12 \\
 & $y_1$ & (99, 61) & (12, 61) & 12 & 61 \\
\midrule
$\mathbf{u}$ & $y_1$ & (99, 61) & (12, 61) & 12 & 12 \\
 & $y_1$ & (70, 55) & (33, 55) & 33 & 33 \\
 & $y_1$ & (89, 71) & (52, 71) & 52 & 52 \\
 & $y_1$ & (95, 64) & (27, 64) & 27 & 27 \\
 & $y_1$ & (52, 33) & (19, 33) & 19 & 33 \\
\midrule
$\mathbf{v}_1$ & $y_1$ & (99, 61) & (12, 61) & 12 & 12 \\
 & $y_1$ & (95, 64) & (27, 64) & 27 & 27 \\
 & $y_1$ & (75, 49) & (24, 49) & 24 & 24 \\
 & $y_1$ & (96, 46) & (20, 46) & 20 & 20 \\
 & $y_1$ & (70, 55) & (33, 55) & 33 & 55 \\
\midrule
$\mathbf{v}_1$, pre-MLP & $y_1$ & (99, 61) & (12, 61) & 12 & 12 \\
 & $y_1$ & (70, 55) & (33, 55) & 33 & 33 \\
 & $y_1$ & (95, 64) & (27, 64) & 27 & 27 \\
 & $y_1$ & (87, 60) & (34, 60) & 34 & 34 \\
 & $y_1$ & (89, 71) & (52, 71) & 52 & 71 \\
\midrule
L13 full at $y_2$ & $y_2$ & (61, 99) & (61, 12) & 12 & 12 \\
 & $y_2$ & (56, 97) & (56, 27) & 27 & 27 \\
 & $y_2$ & (55, 70) & (55, 33) & 33 & 30 \\
 & $y_2$ & (71, 89) & (71, 52) & 52 & 59 \\
 & $y_2$ & (33, 52) & (33, 19) & 19 & 52 \\
\midrule
$\mathbf{v}_2$ & $y_2$ & (61, 99) & (61, 12) & 12 & 12 \\
 & $y_2$ & (71, 89) & (71, 52) & 52 & 52 \\
 & $y_2$ & (44, 90) & (44, 11) & 11 & 111 \\
 & $y_2$ & (45, 79) & (45, 11) & 11 & 111 \\
 & $y_2$ & (55, 70) & (55, 33) & 33 & 55 \\
\bottomrule
\end{tabular}

\end{table}

\begin{table}[!ht]
  \centering
  \small
  \caption{Generated answers, as in Table~\ref{tab:app-whole-examples-a}, under the remaining patches of Figures~\ref{fig:app-whole-shared} and~\ref{fig:app-whole-l15}.}
  \label{tab:app-whole-examples-b}
  \begin{tabular}{llcccc}
\toprule
patch & perturbed & clean $(y_1, y_2)$ & corrupted $(y_1, y_2)$ & $r$ & generated \\
\midrule
$\mathbf{v}_1$ \& $\mathbf{v}_2$ & $y_1$ & (99, 61) & (12, 61) & 12 & 12 \\
 & $y_2$ & (61, 99) & (61, 12) & 12 & 12 \\
 & $y_2$ & (44, 90) & (44, 11) & 11 & 111 \\
 & $y_2$ & (45, 79) & (45, 11) & 11 & 111 \\
 & $y_2$ & (55, 70) & (55, 33) & 33 & 55 \\
\midrule
$(\mathbf{v}_1, \mathbf{v}_2)$ plane & $y_1$ & (99, 61) & (12, 61) & 12 & 12 \\
 & $y_2$ & (61, 99) & (61, 12) & 12 & 12 \\
 & $y_2$ & (62, 94) & (62, 11) & 11 & 111 \\
 & $y_2$ & (56, 88) & (56, 11) & 11 & 111 \\
 & $y_1$ & (89, 71) & (52, 71) & 52 & 71 \\
\midrule
L14 full & $y_2$ & (61, 99) & (61, 12) & 12 & 12 \\
 & $y_1$ & (70, 55) & (33, 55) & 33 & 33 \\
 & $y_2$ & (55, 70) & (55, 33) & 33 & 30 \\
 & $y_2$ & (71, 89) & (71, 52) & 52 & 59 \\
 & $y_1$ & (99, 61) & (12, 61) & 12 & 61 \\
\midrule
L15 DAS & $y_1$ & (99, 61) & (12, 61) & 12 & 12 \\
 & $y_1$ & (70, 55) & (33, 55) & 33 & 33 \\
 & $y_2$ & (62, 75) & (62, 10) & 10 & 102 \\
 & $y_2$ & (55, 83) & (55, 10) & 10 & 105 \\
 & $y_2$ & (61, 99) & (61, 12) & 12 & 61 \\
\bottomrule
\end{tabular}

\end{table}

\FloatBarrier

\end{document}